%% file: main.tex
\documentclass{article} %
\usepackage{iclr2027_conference,times}

\input{math_commands.tex}

\usepackage{hyperref}
\usepackage{url}
\usepackage{booktabs}
\usepackage{graphicx}
\usepackage{amsmath}
\usepackage{amssymb}
\usepackage{xcolor}
\usepackage{float} %

\usepackage[most]{tcolorbox}

\definecolor{cardframe}{RGB}{176,180,196}
\definecolor{cardtitle}{RGB}{236,238,245}
\tcbset{cardstyle/.style={enhanced, colback=white, colframe=cardframe,
  colbacktitle=cardtitle, coltitle=black, boxrule=0.4pt, arc=1.5pt,
  left=3pt, right=3pt, top=2pt, bottom=2pt, toptitle=1pt, bottomtitle=1pt,
  fonttitle=\footnotesize\bfseries, fontupper=\scriptsize, fontlower=\scriptsize}}
\newcommand{\taskid}[1]{{\ttfamily\tiny\color{black!55}#1}\par\smallskip}
\newtcolorbox{taskcard}[2]{cardstyle, title={#1}, before upper={\taskid{#2}}}
\NewDocumentEnvironment{editcard}{m m m +b}{%
  \begin{tcolorbox}[cardstyle, title={#1},
    sidebyside, sidebyside align=top seam, lefthand width=2.4cm,
    sidebyside gap=6pt, segmentation hidden]
  \centering\includegraphics[width=\linewidth,height=2.4cm,keepaspectratio]{#3}
  \tcblower \taskid{#2}#4
  \end{tcolorbox}}{}
\newcommand{\taskprompt}[1]{\par\smallskip{\sffamily #1}}
\usepackage{multicol}
\newcommand{\taskgroup}[1]{\par\smallskip{\footnotesize\bfseries #1}\par\vspace{1pt}}
\newcommand{\taskentry}[4]{{\tiny\par\noindent{\scriptsize\textbf{#1}}\par\nopagebreak
  \noindent{\ttfamily\color{black!55}#2}\hspace{0.5em}#3.\enspace{\sffamily #4}\par}\vspace{2.5pt}}
\NewDocumentEnvironment{editcardv}{m m m +b}{%
  \begin{tcolorbox}[cardstyle, title={#1}, fontupper=\tiny, fonttitle=\scriptsize\bfseries]
  \centering\includegraphics[width=\linewidth,height=1.25cm,keepaspectratio]{#3}\par\smallskip
  \raggedright\taskid{#2}#4
  \end{tcolorbox}}{}
\newcommand{\tasksource}[2]{\par\smallskip\textbf{Source.} #1}
\newcommand{\taskorigin}[1]{\par\smallskip\textbf{Source.} #1}

\newenvironment{promptbox}
  {\par\smallskip\begin{list}{}{\leftmargin=1.2em\rightmargin=0.6em}\item[]%
   \footnotesize\ttfamily\raggedright}
  {\end{list}\smallskip}

\title{PixelBench: Evaluating Language Models \\ for Pixel-Art Creation}

\author{%
Mutian Hong$^{1,3}$ \quad
Jianxuan Yao$^{2,*}$ \quad
Mingkun Yue$^{2,*}$ \\
\textbf{Changkai Li}$^{1}$ \quad
\textbf{Hower Yang} \quad
\textbf{Haibin Ling}$^{1}$ \\
$^1$Westlake University \quad $^2$ShanghaiTech University \quad $^3$Mind Lab \\
{\small $^*$Jianxuan Yao and Mingkun Yue contributed equally.}
}

\iclrfinalcopy %
\begin{document}

\maketitle

\begin{abstract}
Pixel art is widely used in games and digital content creation, requiring
readable forms, coherent spatial relationships, and harmonious colors under
limited resolution and palette constraints. Language models show growing
promise in visual creation, yet their pixel art creation capabilities lack
systematic evaluation. We introduce PixelBench, formalizing text-to-image
generation and instruction-guided editing for pixel art and providing a
benchmark for these tasks. We further present a creation environment in which
language model agents invoke drawing tools through code to generate and edit
pixel art. We assess the quality of generated artwork through human evaluation
and employ a multimodal large language model as a judge, validating the
agreement between its automated assessments and human preferences. We compare
language models and configurations across diverse subjects and canvas sizes.
In our experiments, GPT-6 Astra achieves both the highest average judge score
and the best average human rank, highlighting the potential of general-purpose
language models for pixel art creation. We release the tasks, all artworks,
the human rankings, and the judge protocol.
\end{abstract}

\begin{figure}[H]
\centering
\includegraphics[width=\textwidth]{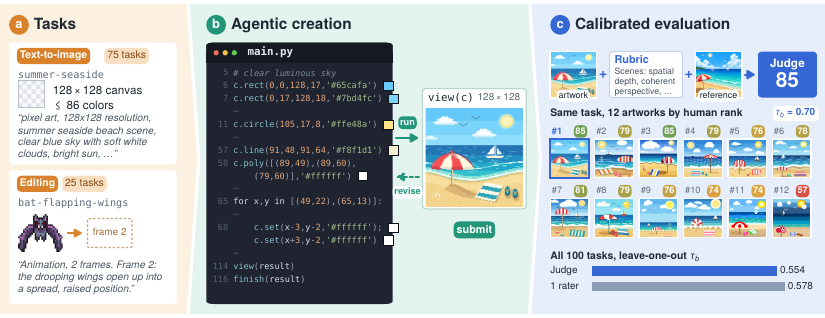}
\par %
\caption{PixelBench on the task \texttt{summer-seaside}. (a) A task gives a
written requirement and a native canvas; an editing task also gives the artwork
to modify. (b) The model writes, runs, views, and revises a drawing program
until it submits. The code shown is part of the program submitted by GPT-6 Astra
in the evaluation. (c) The calibrated judge scores each artwork given a rubric and a
reference image. Below: the task's 12 submissions by mean rank over the ten
raters, \#1 to \#12 reading left to right and top to bottom, each with its judge
score; and leave-one-out agreement over all 100 tasks, for the judge and for
the average held-out rater.}
\label{fig:framework}
\end{figure}

\section{Introduction}

Pixel art is digital painting in which the individual pixel is the smallest
unit of drawing. It is one of the most common art forms in two-dimensional
games, used for characters and their animation frames, props, map tiles, and
interface elements, and it remains in wide use in independent and commercial
games today \citep{silber2015pixelart}. Beyond games it appears in application
icons, emoji, and avatars, and is carried over into physical media such as
fuse beads and cross-stitch. Pixel artists work in dedicated editors, among
them Aseprite, Pyxel Edit, GraphicsGale, and Piskel
\citep{aseprite2026,pyxeledit2026,graphicsgale2026,piskel2026}. These tools are built
around how the medium is made: the artist places colors cell by cell with a
pencil tool on a magnified native canvas, maintains a limited palette,
organizes the picture in layers, builds animation frame by frame, and exports
a sprite sheet laid out on a fixed grid. The constraints behind that workflow
are what make the medium hard: the canvas is the size of the finished work, so
a $32\times32$ character has only 1024 cells \citep{wu2022myos}, and because
the work is displayed at integer nearest-neighbor magnification without
interpolation, every pixel stays a hard-edged block
\citep{saravanan2022pixelvqvae}. The color of every cell is therefore a design
decision that directly determines whether a silhouette is clean and whether a
form holds together.

\newcommand{\teaserpanel}[3]{%
  \makebox[0.29\textwidth]{%
  \begin{tabular}[t]{@{}c@{}}
  \includegraphics[height=3.3cm]{figures/fig_sailboat_#1.png}\\[1pt]
  \small #2\\[-1pt]
  {\scriptsize\color{black!65} #3}
  \end{tabular}}}
\begin{figure}[t]
\centering
\hspace*{0.06\textwidth}%
\teaserpanel{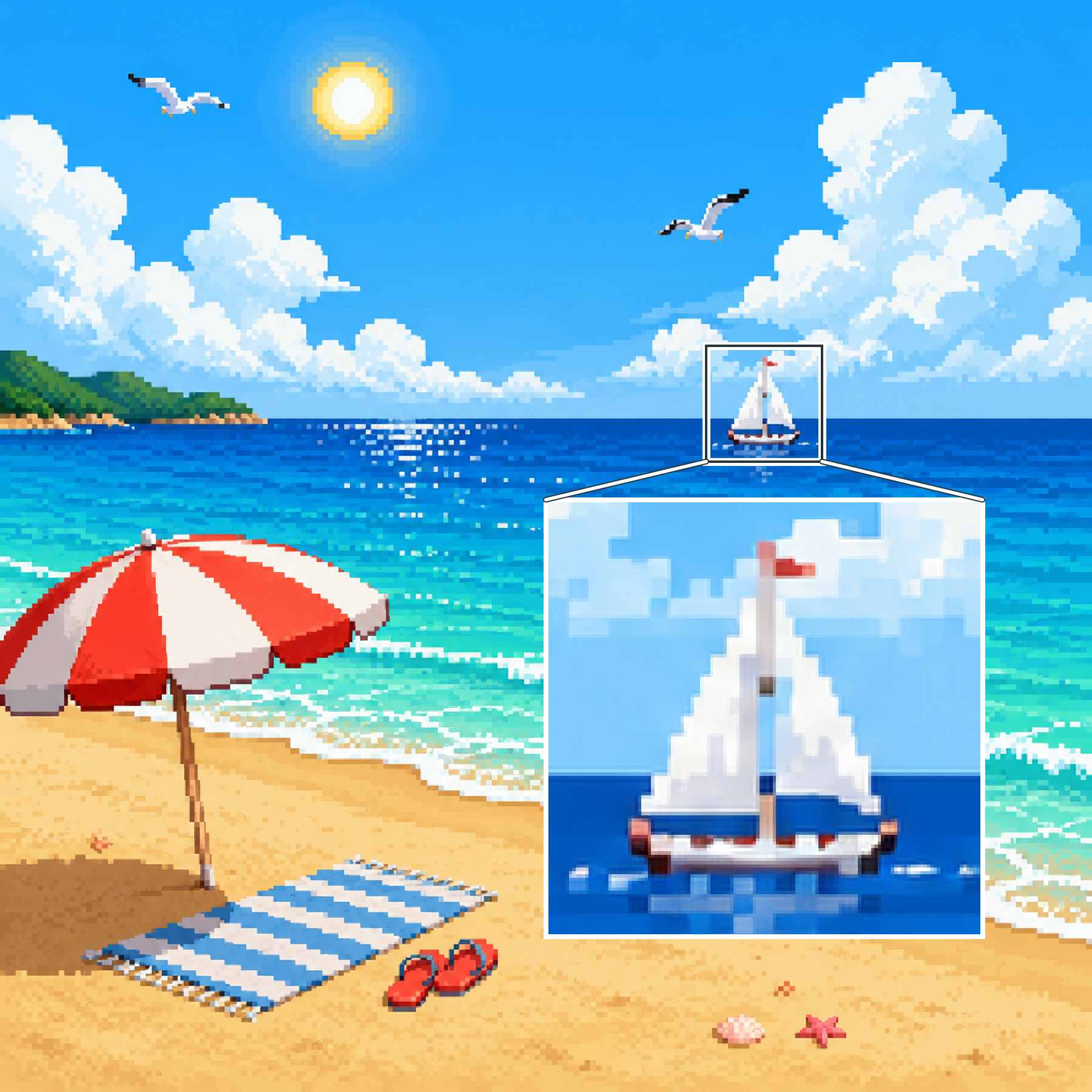}{(a) GPT-Image-2}{$1254\times1254$ $\cdot$ 234,693 colors}\hfill
\teaserpanel{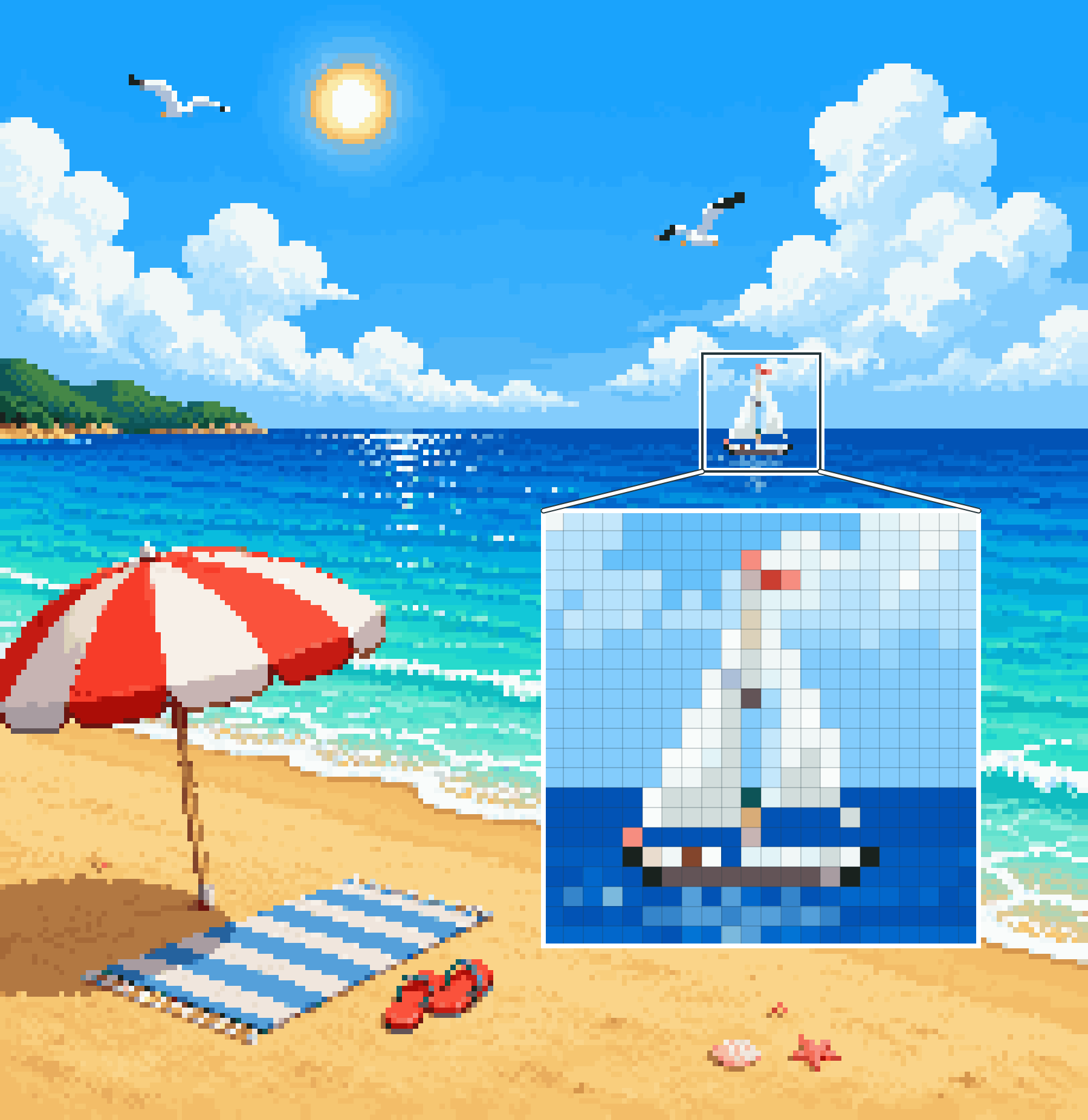}{(b) + Pixel Snapper}{$203\times209$ $\cdot$ 86 colors}\hfill
\teaserpanel{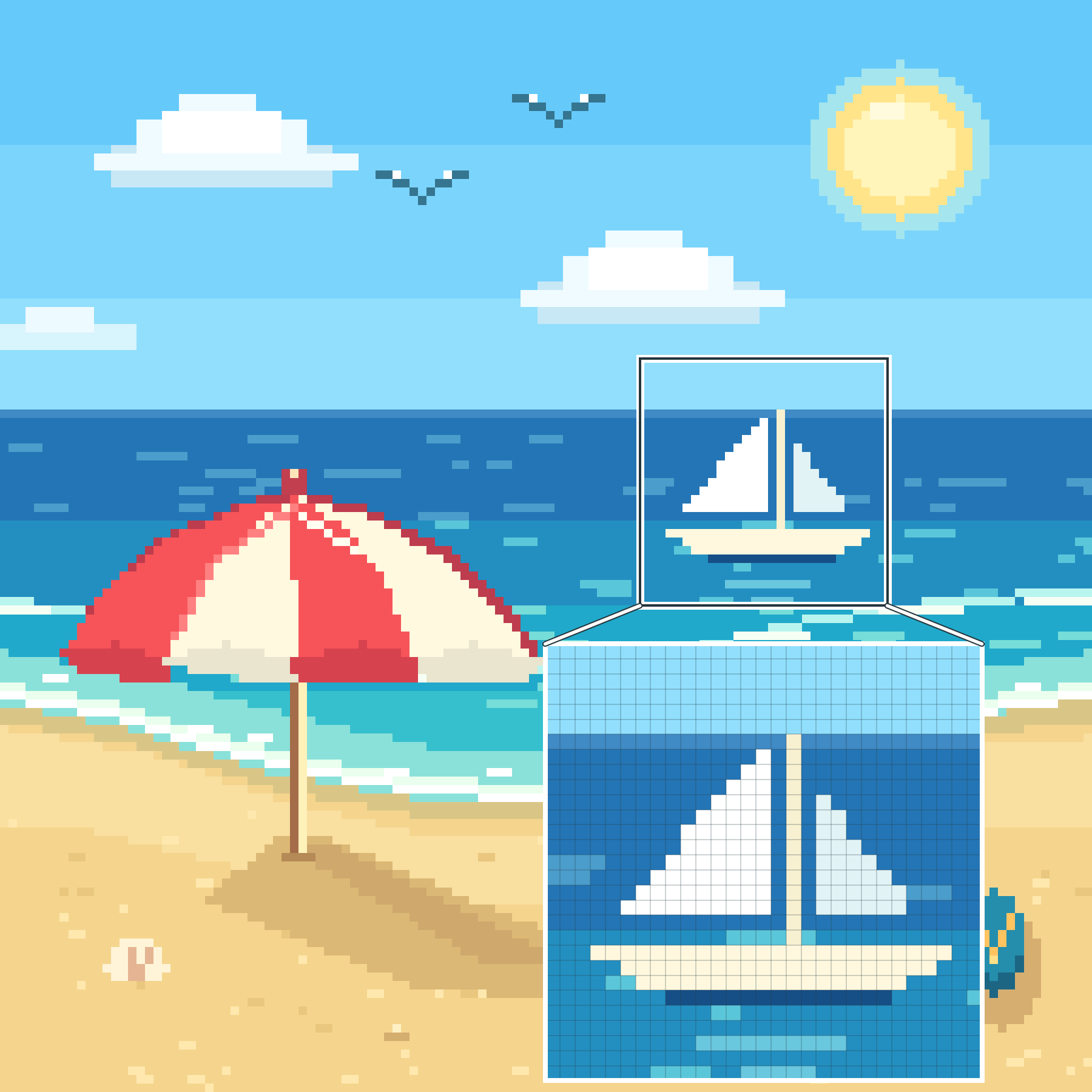}{(c) GPT-6 Astra}{$128\times128$ $\cdot$ 63 colors}%
\hspace*{0.06\textwidth}
\caption{An example PixelBench task, \texttt{summer-seaside}: a $128\times128$ summer seaside with at most 86
colors; insets enlarge the sailboat. (a) The output of an image generation
model imitates pixel art, but its pixels are not clean cells: many are not
square, straddle cell boundaries, or blend several colors, and no fixed palette
constrains them. (b) Pixel Snapper \citep{pixelsnapper2025} infers a grid from
that image and quantizes it to the task's 86 colors; every cell is now
uniform, but the grid is $203\times209$ rather than the required canvas. (c) GPT-6 Astra, working in
PixelBench, writes a program that draws on the native canvas cell by cell;
shown is its benchmark submission.}
\label{fig:grid}
\end{figure}

Because drawing by hand is expensive, several technical routes aim to have
models produce pixel art. The first uses image generation models directly.
These models produce continuous images at a fixed high resolution, latent
diffusion models for example through a latent space and a decoder
\citep{rombach2022ldm}; with a style prompt they produce pixel-style pictures,
but those pictures do not lie on a pixel grid. The raw GPT-Image-2 output for
the $128\times128$ \texttt{summer-seaside} task of Figure~\ref{fig:grid} carries
234,693 distinct colors, and its apparent blocks are neither uniform in color
nor aligned to any single pitch. Tools exist that recover a grid from such a
picture and quantize its palette \citep{pixelsnapper2025}, but they can only
infer the grid the picture happens to imply, and that grid is a property of
the generated image rather than of the requirement: on this image it comes out
$203\times209$, not the $128\times128$ the task asks for. The second route is
pixelization, which converts an existing ordinary image into pixel art
\citep{gerstner2012pia,han2018dup,wu2022myos}. In both routes the work is
first produced as a continuous bitmap and the grid and palette are imposed
afterwards; the per-cell color decisions are made by a conversion algorithm
rather than by the creator, and the output carries no layers or other editable
structure.

Language models offer another route: drawing directly on a discrete canvas at
native resolution, as they already do for vector graphics and small drawing
programs
\citep{rodriguez2023starvector,yang2025omnisvg,rodriguez2026vectorgym,nedungadi2026mosaics,pixelartbench2026}.
Pixel art suits this
form of expression: the canvas is an integer matrix, colors form a finite
symbolic set, and drawing operations are discrete geometry that a program
expresses directly. What this route yields also has different properties: the
creator makes the per-cell decisions, any single pixel can be located and
changed exactly, and the creation process persists as a program with its
layers intact, one that can be re-run and edited further and so enters a human
workflow directly.

What language models can actually do on this route has not been measured
credibly. Existing evaluations generate on small $24\times24$ grids in one
turn, without tools or revision, and their automatic judges are not
calibrated against complete human rankings
\citep{pixelartbench2026,nedungadi2026mosaics}.
Judging pixel-art quality depends heavily on human perception: whether a
silhouette flows, whether volume holds, whether a palette coheres are not
properties a rule can compute. Evaluating pixel-art creation by language
models therefore requires solving two problems at once: how to define tasks at
production-relevant canvas sizes and have models complete them, and how to
build an evaluation protocol that is both grounded in human judgment and able
to run at scale.

We propose PixelBench to answer both, as Figure~\ref{fig:framework} lays out.
On tasks, PixelBench contains 75
text-to-image problems and 25 editing problems. The generation problems sit on
native canvases from $16\times16$ to $256\times256$ and cover seven subjects:
characters, creatures, scenes and architecture, objects, icons and interface
elements, materials, and effects. The editing problems come in three types:
instruction editing, animation, and multiview. On method, we provide an
agentic creation environment: the model writes a Python program against a drawing
API, runs it, views the rendered result, edits the source, and runs again until
it submits. On evaluation, we calibrate a multimodal judge that scores
one artwork at a time, on three fronts: how the artwork is presented, the
rubric, and a same-task reference image. The judge was settled on earlier
artworks; ten raters with professional art backgrounds then ranked all 1,200
artworks of the evaluation, an out-of-sample test of the judge. The judge's order of a
task's artworks agrees with the raters' consensus at a per-task Kendall
$\tau_b$ of $0.557$. To make this number interpretable, we put
the judge and the raters on the same footing: we hold out one rater at a time
and measure how well that rater, and the judge, agree with the mean ranking of
the other nine. The ten raters reach $0.465$ to $0.649$ this way and the
judge $0.554$, inside the range the raters themselves span and above three of
them.
At the model level the judge and human leaderboards correlate at Spearman
$\rho = 0.958$.

Under this protocol we evaluate 12 frontier model configurations. GPT-6 Astra
attains the highest judge score and the best mean human rank on both
generation and editing, and nine of the ten raters individually rank it
first. Raters agree most, and models separate most, on characters and on
larger canvases. With the judge we also score settings no rater ranked
(Section~\ref{sec:variants}): repeated creation, reasoning effort, a
direct-answer baseline without tools, and image observation, whose removal
lowers GPT-6 Astra's editing score from $88.0$ to $77.8$ while generation
barely moves: an edit starts from an artwork someone else drew, and seeing it
lets the model understand it better.

The contributions of this paper are as follows.

\begin{enumerate}
\itemsep0em
\item \textbf{We build PixelBench, which formalizes pixel-art generation and
editing for language model agents.} An artwork is a color-index matrix at a
specified native canvas size; the environment guarantees canvas size and color
ceiling, and form, composition, and color are left to human judgment.
\item \textbf{We release a pixel-art creation environment for language model
agents.} The model writes a program that calls drawing tools and creates
through a loop of running, viewing, and revising its source, so an artwork is
fully determined by a program that can be re-run, audited, and edited further.
Every configuration in this paper creates in this one environment, and it also
ships as a reusable agent skill.
\item \textbf{We build a multimodal judge for pixel art and calibrate it
against human rankings.} Using complete rankings of 1,200 artworks by ten
raters, we measure how presentation, rubric, and reference image each
change agreement with the consensus order, and, under one leave-one-out
reference, place the judge inside the range the ten raters themselves
span, overall and on every subject, canvas size, and editing type, and above
all ten on the largest canvases.
\item \textbf{We evaluate 12 frontier model configurations and analyze their
behavior.} We report judge and human leaderboards for generation and editing,
and analyze how the differences between models are distributed across subjects,
canvas sizes, and editing types; with the judge we further measure
repeated creation, reasoning effort, a direct-answer baseline without tools,
image observation, and cost.
\end{enumerate}

\section{Related Work}

\textbf{Pixel-art generation and pixelization.}
Dedicated generators for pixel art include GAN-based character sprite
generation across orientations \citep{coutinho2022spritegan} and VQ-VAE
representations tailored to the medium \citep{saravanan2022pixelvqvae}.
Pixelization converts an ordinary image into
pixel art, through superpixel abstraction with a jointly optimized small
palette \citep{gerstner2012pia}, unsupervised domain translation that learns
grid structure and a crisp pixel style \citep{han2018dup}, or staged methods
that let the pixel cell be controlled \citep{wu2022myos}; related tools align
the pixel-style output of an image generation model to a grid and quantize its
palette \citep{pixelsnapper2025}. In all of these the per-cell colors are
decided by a conversion or generation algorithm. PixelBench studies a different mode of creation: a general-purpose language
model, given an open written requirement, starts from an empty canvas or an
existing artwork and makes the per-cell color decisions itself on the native
grid.

\textbf{Program-based drawing.}
Generating vector graphics with language and multimodal models is now
systematically studied, covering image- and text-to-SVG generation and
multi-task evaluation of understanding, editing, and generation
\citep{rodriguez2023starvector,yang2025omnisvg,rodriguez2026vectorgym}:
constructing visual content through code is a viable route. Their
evaluation scores reconstruction and editing against a target image or SVG,
and open-ended text-to-SVG generation with CLIP-style metrics or multimodal
judges that mainly measure semantic alignment
\citep{rodriguez2026vectorgym,cipriano2026svgscore}. On discrete
canvases, Pixel Art Bench \citep{pixelartbench2026} has text
models emit a $24\times24$ palette-indexed grid in one turn and computes
Elo with a pairwise judge, and Autoregressive Mosaics
\citep{nedungadi2026mosaics} has models write short programs that call drawing
primitives on a $24\times24$ canvas in order to probe two-dimensional spatial
reasoning; DrawingBench \citep{kim2025drawingbench} instead evaluates drawing
through mouse-level GUI actions. PixelBench evaluates complete artworks and
edits on canvases
up to $256\times256$, created through a loop in which the model operates an
explicit integer canvas with discrete primitives and revises its own program,
and calibrates its judge against complete human rankings.

\textbf{Human-aligned evaluation of visual generation.}
Using a strong model as a judge has become standard practice for open-ended
generation \citep{zheng2023mtbench}, and its credibility depends on calibration
against human preference \citep{chiang2024arena}. In the visual domain, GenEval
and T2I-CompBench++ decompose text-image alignment into checkable factors
\citep{ghosh2023geneval,huang2023t2icompbench}, while SVGauge
and SVG-Score study human-aligned evaluation for SVG generation
\citep{zini2025svgauge,cipriano2026svgscore}. ImagenWorld compares each
annotator and a VLM judge with the mean of the remaining annotators
\citep{sani2026imagenworld}, the leave-one-out reference we also use, and
K-Sort Eval corrects a VLM judge with human rankings of $K$ outputs at a time
\citep{li2026ksorteval}. For pixel art, where a task has no ground-truth
answer, we calibrate a judge of overall quality, visual quality together with
instruction following, against complete per-task rankings by ten raters.

\section{PixelBench}
\label{sec:bench}

Each PixelBench task gives a written requirement and a specification, and the
model delivers a pixel artwork $A = (G, P, M)$, where
$G \in \{0,\dots,|P|-1\}^{H\times W}$ is a matrix of color indices on the
native canvas, $P$ is the palette, and $M$ records the native size, the
background policy, and the frame structure. Stored in this form, an artwork's
color count, modified regions, and inter-frame differences are exactly
computable. An artwork is rendered as an image only to be observed or
evaluated, by nearest-neighbor magnification without interpolation; before it
reaches the judge, a transparent margin is added on its left and top, a change
whose effect on judge--human agreement Section~\ref{sec:calib} measures.

The environment guarantees the specification. An artwork of the wrong size or
with too many colors raises an error as soon as the program declares it, and
the model fixes the program and runs it again. The specification constrains size and color,
not style: whether to use dithering, outlines, or hand-placed transitional
colors is the creator's choice, and meeting the specification does not make an
artwork good pixel art. Good pixel art reads clearly at native size, has clean silhouettes,
groups its colors by light and shade to model form and volume, has a coherent
composition and palette, and delivers what the task asks for; these criteria
resist being written down as computable rules and are judged by human ranking
and by the calibrated judge (Section~\ref{sec:eval}).

Every one of the 100 tasks was written and reviewed by people, with generative
AI assisting on descriptive detail. Appendix~\ref{app:tasks} gives
a card for every task with its statement, canvas, and color ceiling, and, for
editing tasks, the source artwork with a citation of the work it comes from;
Appendix~\ref{app:gallery} shows every artwork produced for every task.

\subsection{Text-to-image generation}
Each of the 75 generation tasks gives an English requirement and a native
canvas size, and starts from an empty canvas. The tasks cover seven subjects:
\emph{objects} (21), \emph{scenes and architecture} (16), \emph{characters}
(13), \emph{creatures} (10), \emph{icons and interface elements} (7),
\emph{materials} (5), and \emph{effects} (3). Canvases
range from $16\times16$ to $256\times256$, most often $64\times64$ (26 tasks)
and $32\times32$ (18).

Every task also carries a color ceiling, and the ceiling plays two different
roles. On seven tasks it is a tight budget of four to ten colors and it binds:
these are $16\times16$ and $32\times32$ pieces where working within a handful
of colors is part of the task, and on each of them the most colorful submission
uses the full allowance. Everywhere else it is
set generously, at 86 colors for the bulk of the benchmark and 120 to 140 for
the four $256\times256$ tasks, and models stay far below it, the median
submission using 19 colors against a ceiling of 86; its role there is to keep a
continuous-tone output from passing as pixel art.

\subsection{Editing}
The 25 editing tasks each give an existing pixel artwork and an instruction,
and require the modification to be made while the original is preserved. They
come in three types. The 13 \emph{instruction editing} tasks produce a single
frame, for example melting an ice cream or removing a character from a scene and
filling in the background. The 10 \emph{animation} tasks take the original
as the first frame and continue a specified action, such as a punch or a flag
waving, seven of them as two-frame and three as three-frame
sequences. The 2 \emph{multiview} tasks ask for a three-view turnaround of a character
and for two rotated views of a house. Source canvases range from
$27\times22$ to $256\times128$. Editing tasks set no color ceiling, and the
model may enlarge the canvas as the edit requires; all frames of a multi-frame
result share one size.
The originals were collected from the internet, and three were drawn by the
authors; sources are listed in Appendix~\ref{app:tasks}. Editing tests
a capability distinct from generation: understanding the structure of an
existing artwork, changing only the part that should change, and keeping the
rest consistent in content and style.

\section{Agentic Pixel-Art Creation}
\label{sec:environment}

The creation environment places every model under identical conditions. The
system prompt documents the file protocol and the drawing API, and carries no
pixel-art technique guidance and no exemplar artwork; every model receives the
same prompt, tools, and budget, and composition, color use, and drawing order
are left entirely to the model. The leaderboard therefore compares the models
themselves as pixel-art creators, not prompt designs tailored to any one of
them.

\textbf{File protocol.} The model works in a source workspace and delivers
the artwork as a Python program. It manages source through
\texttt{write\_file}, \texttt{edit\_file}, \texttt{read\_file}, and
\texttt{list\_files}, and executes it through \texttt{run(path)}. Every run
starts from the task input in a fresh subprocess, and no variable or canvas
survives from the previous run, so an artwork is fully determined by its
source. The subprocess runs in a sandbox: it can read and write only its
own run directory, has no network access, and cannot start other programs.
A program declares its output with \texttt{finish(canvas)} or
\texttt{finish([f1, f2, \dots])}, at which point the environment checks the
specification (Section~\ref{sec:bench}) and freezes the result; only
\texttt{run(path, submit=True)} submits formally and ends the task. The budget
counts source modifications: 25 \texttt{write\_file}/\texttt{edit\_file}
calls for generation and 15 for editing, with reads and runs free. At the last
modification the environment warns the model to submit.

\textbf{Drawing API.} Programs have a pixel-art drawing library preloaded.
\texttt{new\_canvas(W, H)} creates a canvas; several canvases can serve as
layers and are composited in order by \texttt{flatten}. A canvas provides
single pixels, lines, rectangles, circles and ellipses, polygons, SVG paths
with B\'ezier curves and arcs, flood fill, mirroring, shifting, outlines,
checkerboard and Bayer dithering, and text. No primitive anti-aliases, and
\texttt{None} erases. Colors can be given by name, hex, or RGBA and registered as named
colors that form the palette of the artwork. \texttt{select(x, y, w, h)} lifts
a region that can be flipped, resized, or rotated (nearest-neighbor or
RotSprite) and then copied or moved elsewhere; in editing and animation tasks
this is how models move limbs and reuse parts. An artwork can also be saved as
a layered \texttt{.ase} or \texttt{.ora} document and opened for further work in
pixel-art editors such as Aseprite \citep{aseprite2026}. Every primitive
lands on the native grid without anti-aliasing, so the deliverable remains a
native pixel matrix; Appendix~\ref{app:svg} compares it with SVG generation.

\textbf{Creation loop.} Figure~\ref{fig:framework}b shows the loop.
\texttt{view(canvas)} magnifies the canvas with nearest-neighbor sampling to
about 512 pixels on the long edge and returns it to the model as an image, and
it is the only channel through which a model obtains an image observation: for
the five evaluated models without image input, and for GPT-6 Astra when
image observation is switched off (Section~\ref{sec:variants}), \texttt{view}
returns no image and the model
reads the artwork as a character picture instead
(Appendix~\ref{app:textobs}). The environment records the source snapshot of
every run, standard output, canvas checkpoints, and every preview the model
requested. Editing tasks provide the source artwork as
\texttt{input\_canvas}, and these records show how models work from it: on
all 25 editing tasks, GPT-6 Astra first writes a short inspection script
that magnifies the original and prints its character matrix and palette, and
only then writes the main program that modifies it. On \texttt{fist-swing},
for instance, GPT-6 Astra inspects the $29\times44$ original, widens the canvas
to $36\times46$ so that the punch has room, and places the original
unchanged in the first frame with \texttt{blit}. It copies the original,
clears the rows holding the raised forearm, and redraws the arm extended as
polygons in the original's outline, skin, and shading colors; it then views
both frames and submits. Across the 25 editing tasks, 13 of GPT-6 Astra's
submissions enlarge the canvas and 12 contain two or three frames. Blitting, clearing, and
selections let a model keep the original's pixels wherever the edit does not
reach, and the preview after each run is what tells it whether the change reads
at native size.

\textbf{Agent skill.} The environment also ships as an agent skill, a folder of
instructions, scripts, and resources an agent loads when a task calls for them
\citep{anthropic2025agentskills}, bringing the same canvas representation and
drawing primitives to general-purpose coding agents.

\section{Evaluation Protocol}
\label{sec:eval}

\subsection{Human ranking}
\label{sec:human}

Ten external raters ranked all 1,200 artworks (100 tasks $\times$ 12 model
configurations). All ten have professional art backgrounds, mainly in art and
game design, with experience in graphic design, years of painting, or
pixel-style game art. For each task a rater saw the task statement, the canvas
requirement, and the 12 artworks, anonymized under random labels and shown
in a random initial order drawn for each rater, and ranked
them from best to worst on visual quality and on how well each follows the
instruction. Eight raters also saw a reference for what the task asks for:
for generation a GPT-Image-2~\citep{openai2026gptimage2} rendering of the same
statement, and for editing the original plus the reference image the judge
receives in Section~\ref{sec:calib}; the other two ranked every task
without a reference, seeing only the original for editing. The frames of each animation were
played in a loop so that raters could judge whether the motion is continuous.
The interface recorded each rater's initial order and time per task, and
the initial order has little effect on the final ranking
(Appendix~\ref{app:raters}, which also shows the interface).
Across the 1,000 ranking lists, the per-task mean Kendall
$\tau_b$ between pairs of raters is $0.459$. An artwork's ten ranks average
to its \emph{mean human rank}; a task's twelve form the \emph{consensus order}
the judge is calibrated against, and averaging over tasks gives a
configuration's place on the \emph{human leaderboard}.

\subsection{The multimodal judge}
\label{sec:judge}

The automatic judge serves evaluation at scale and lets the community compare
new models on PixelBench; the judge leaderboard is therefore the
main table of this paper, with the human leaderboard supplying its
calibration. The judge scores a single artwork independently: each call
contains only the task statement, the canvas and color requirement, and one
anonymized artwork, and outputs a single integer from 0 to 100. No two calls
share context, and no competing candidate is ever shown.
An editing call also shows the original, a multi-frame result is
laid out in order as one image with the prompt stating the layout, and a
material tile is shown repeated $2\times2$ to expose its seams.
The judge model is GPT-6 Astra with reasoning effort
set to \texttt{xhigh}, chosen for its multimodal understanding: it scores
87\% on MMMU-Pro \citep{yue2025mmmupro}, among the highest scores an
independent tracker reports \citep{aa2026mmmupro}. Appendix~\ref{app:prompts}
gives the complete prompts. Scoring a new model's 100 submissions therefore
costs 100 calls to the judge model, and scoring each artwork once is enough for
a leaderboard: 62 of
the 66 pairs of configurations in Table~\ref{tab:main} keep their order in at
least $99\%$ of simulated repeat evaluations, and GPT-6 Astra's margin over
every other configuration holds in all of them (Appendix~\ref{app:repeat}).
The randomness of creation itself is measured separately, by creating the
same tasks again (Section~\ref{sec:variants}).

We measure agreement as the per-task Kendall $\tau_b$
\citep{kendall1945tau} between the judge's twelve scores for a task and the
negated mean human ranks, so that both run best to worst and $\tau_b = 1$ means
the judge reproduces the consensus order; agreement is
averaged over tasks with equal weight, with 95\% confidence
intervals from 10{,}000 bootstrap resamples at the task level. To compare two
judge protocols, or two configurations of one model, we take their
difference on every task and report the mean of this paired difference with
its 95\% interval, which excludes zero when the two differ significantly at
this sample size.

\subsection{Calibrating the judge}
\label{sec:calib}

\begin{table}[t]
\caption{Calibrating the judge. Per-task mean Kendall $\tau_b$ against the
consensus order of the ten raters, with a 95\% bootstrap interval
over tasks. Each row adds one change to the row above, and
Appendix~\ref{app:prompts} gives the exact prompt of each version; the last
row is the final protocol.}
\label{tab:calib}
\centering
\small
\setlength{\tabcolsep}{4pt}
\begin{tabular}{@{}llcccc@{}}
\toprule
Protocol & Change & All (100) & 95\% CI & Generation (75) & Editing (25) \\
\midrule
Simplest       & single image, overall score & 0.528 & [0.484, 0.572] & 0.528 & 0.529 \\
+ presentation & 1/16 transparent margin     & 0.532 & [0.497, 0.568] & 0.531 & 0.536 \\
+ rubric       & by subject and task type    & 0.544 & [0.506, 0.581] & 0.553 & 0.517 \\
+ reference    & same-task reference image   & \textbf{0.557} & [0.520, 0.593] & \textbf{0.563} & \textbf{0.538} \\
\bottomrule
\end{tabular}
\end{table}

Starting from the simplest protocol, we added three changes in turn. The
prompts were settled on a development subset of earlier artworks, against
rankings by members of the research team (Appendix~\ref{app:prompts}), and
then frozen. Table~\ref{tab:calib} runs each version on all 100 tasks of the
evaluation and checks it against the ten raters, so every number in
this section tests the protocol on artworks and raters it has never
seen. In the simplest protocol the artwork is magnified with
nearest-neighbor scaling so that its longer side is $512$ pixels and the prompt
asks only for an overall quality score, which reaches a per-task mean $\tau_b$
of $0.528$.

\textbf{Presentation.} We add a transparent margin of one sixteenth of the
canvas dimension on the left and top of the artwork before magnifying.
On a small canvas, magnifying to $512$ turns every pixel into a block of
$16\times16$ or larger that a patch of a patch-based vision encoder can fall
entirely inside; offsetting the grid by a fraction of a block breaks that
alignment. Agreement moves to $0.532$.

\textbf{Rubric.} The rubric reaches the judge as prompt text: every call
appends a supplement for the task's category after the task statement. Each
generation subject has its own supplement naming what to weigh: pose,
silhouette, and anatomy for characters; spatial depth and the separation of
foreground from background for scenes; continuity across the tile boundaries
for materials. Four of the seven subjects append one further sentence chosen by
canvas size, asking that every pixel be placed deliberately at 32 pixels and
below, and that detail serve form above it. The editing supplement weighs
instruction compliance, preservation of what the instruction does not change,
and craft; animation adds inter-frame continuity and pose plausibility, and
multiview adds identity across views. Agreement rises to $0.544$.

\textbf{Reference image.} Finally, we show the judge a same-task reference
image before the candidate. For generation it is GPT-Image-2's output for the
same statement, and the prompt states that it is not pixel art, not ground
truth, and not to be scored against for resemblance. Among the editing
tasks, animation, multiview, and one instruction-editing task use the
corresponding frames from the asset the original comes from, and the other
twelve use GPT-Image-2's edit. Agreement reaches $0.557$ $[0.520, 0.593]$,
the highest of the four versions, on generation and on editing alike. The raters ranked
with these same references in view (Section~\ref{sec:human}), so both now work
from a shared point of comparison, and the judge--human agreement we
report is that of this shared-reference protocol. Without a reference the
human leaderboard barely moves: the two raters who never saw one produce a
leaderboard that correlates with the other eight's at Spearman $\rho = 0.972$
(Appendix~\ref{app:raters}). We fix the protocol here; every judge result
below comes from this version.

\textbf{The judge falls within the spread of the raters.} What decides a
leaderboard is the order of the twelve artworks of a task, so that is where we
place the judge and the raters side by side, using leave-one-out: hold out one
rater, compare both that rater and the judge against the mean rank of the other
nine, and rotate through all ten. The held-out raters reach $0.465$
to $0.649$, averaging $0.578$ $[0.555, 0.600]$, and the judge reaches
$0.554$ $[0.518, 0.589]$: the judge sits inside the range the raters span,
and above three of them. The per-task paired difference between the judge and
a held-out rater is $-0.023$ $[-0.056, +0.008]$; even at the far end of this
interval, the judge trails the average rater by half as much as the lowest
rater does ($0.113$). The same holds slice by slice: on every subject, canvas
size, and editing type the judge falls within the range of the ten raters, and
on canvases above 64 pixels above all of them (Appendix~\ref{app:slices}),
including materials, where the raters agree least among themselves. At the
model level the judge and human leaderboards correlate at Spearman
$\rho = 0.958$ $(p = 9.5\times10^{-7})$, close to the $0.980$ between two
random halves of the raters.

\textbf{Other judge models agree less with the raters.} Under the same
protocol we also scored the artworks with Claude Opus 5, at reasoning effort
low, and with the open-weight DeepSeek V4.1 Flash, at high. GPT-6 Astra's
agreement with the consensus order exceeds theirs by $0.058$ $[0.016, 0.098]$
and $0.144$ $[0.108, 0.179]$, paired over tasks, and under leave-one-out
Claude Opus 5 falls above one of the ten raters and DeepSeek V4.1 Flash below
all of them. Both still place GPT-6 Astra first, and their leaderboards
correlate with the human one at Spearman $\rho = 0.825$ and $0.958$
(Appendix~\ref{app:crossjudge}).

\section{Results}

Table~\ref{tab:main} gives the results for the 12 model configurations, over
all 100 tasks and separately over generation and editing. GPT-6 Astra ranks
first on both leaderboards, with a judge score of $83.76$ and a mean human
rank of $3.34$; second place goes to Claude Opus 5 on the judge leaderboard
($78.49$) and to Gemini 3.8 Flash on the human one ($4.27$). No configuration
moves by more than two places between the two leaderboards. Gaps of a
point or two, as in the middle of the table, are within what a second
creation run can change (Section~\ref{sec:variants}), so we read the order only
across larger gaps.

On the within-task rank scale of Table~\ref{tab:main}, ties sharing the
mean of their positions, the judge's mean rank differs from the raters' by
$0.52$ places on average.

The models are GPT-6 Astra~\citep{openai2026gpt6astra}, GPT-5.6
Sol~\citep{openai2026gpt56sol}, Qwen3.8 Max~\citep{qwen2026qwen38max}, Claude
Opus 5~\citep{anthropic2026opus5}, Gemini 3.8
Flash~\citep{google2026gemini38flash}, Kimi K3~\citep{kimi2026k3}, DeepSeek V4
Flash, V4.1 Flash, and V4 Pro~\citep{deepseek2026v4}, GLM-5.3
Flash~\citep{zhipu2026glm53flash}, Gemini 3.1 Pro
Preview~\citep{google2026gemini31pro}, and
Qwen3.5-397B-A17B~\citep{qwen2026qwen35}; the DeepSeek snapshots are
\texttt{0731} for V4 Flash and \texttt{0813} for V4 Pro.
Appendix~\ref{app:runs} describes the evaluation runs.

\begin{table}[t]
\caption{Leaderboard over 12 configurations, ordered by mean judge score over
all 100 tasks under the final protocol. The two
rank columns are mean within-task ranks over the twelve configurations, the
judge's from its scores and the human one over the ten raters; both
are lower-is-better and live on the same scale. The value in parentheses is
the reasoning effort every call ran at; the two Qwen models take a thinking
budget instead.}
\label{tab:main}
\centering
\small
\setlength{\tabcolsep}{2pt}
\input{tables/table_main_b2}
\end{table}

\subsection{Generation}
\label{sec:slices}

The gap between models is not equally measurable on every task. Broken down by
subject and canvas size, the 75 generation tasks separate the models most on
large canvases and concrete subjects, through three quantities that move
together (Table~\ref{tab:slices} in Appendix~\ref{app:slices} gives every
slice).

We measure how much a task's twelve artworks differ by the standard deviation
of their judge scores, and a large canvas leaves room for form, depth, and
detail, so a strong and a weak artwork come apart: that spread is $11.5$ on
canvases above 64 pixels and $10.4$ on characters, against $7.8$ on canvases of
16 pixels and below and $7.1$ on objects. Rater agreement follows the same
slices, $0.504$ on the largest canvases and $0.492$ on characters against
$0.343$ on the smallest and $0.263$ on materials. The twelve configurations'
mean human ranks then spread further on the same slices, at $2.38$ on
characters and $2.28$ on the largest canvases against $1.67$ and $1.61$ on the
smallest canvases and materials, and judge--human agreement is higher there
too, $0.609$ and $0.697$ against $0.472$ and $0.460$. This spread is not
independent of rater agreement, since ranks averaged over disagreeing raters
drift toward the middle.

\textbf{The slices that separate the models are also the hardest.} Characters
and creatures carry the strongest structural constraints in the medium: viewers
hold extremely sensitive priors about proportion, pose, and face. Averaged
over the twelve configurations, characters, creatures, and effects score
$67$--$69$, against $78$--$79$ on objects and on icons and interface elements,
and even the best artwork of each task averages $81.5$ on characters against
$88.3$ on icons and interface elements.

\textbf{Small canvases are not easy.} On a $16\times16$ canvas a single
misplaced pixel changes a silhouette, and on the ten tasks of 16 pixels and
below most artworks would need retouching before use. Raters then have little
basis for ordering them, and the human first place goes to seven different
configurations.

\subsection{Editing}
\label{sec:edit}

\textbf{Leaderboard.} On the 25 editing tasks of Table~\ref{tab:main},
GPT-6 Astra again ranks first, with a judge
score of $88.00$ and a mean human rank of $3.74$. The ordering on editing is
related to, but not the same as, the ordering on generation: across the 12
configurations, mean human rank on generation and on editing correlate at
Spearman $0.895$. Claude Opus 5 and Gemini 3.1 Pro Preview rise on the
editing human leaderboard, from fourth to second ($4.25$) and from ninth to
fifth ($5.76$), while Gemini 3.8 Flash drops from second on generation to
fourth ($5.62$).

\subsection{Analyses scored by the judge}
\label{sec:variants}

\begin{table}[t]
\caption{Judge-scored variants, each run on all 100 tasks and scored once by
the judge. The rows marked Table~\ref{tab:main} are the main-table
setting; creation runs 2 and 3 repeat that setting from scratch.}
\label{tab:variants}
\centering
\footnotesize
\setlength{\tabcolsep}{5pt}
\renewcommand{\arraystretch}{0.92}
\input{tables/table_variants_b2}
\end{table}

The judge lets us score settings that no rater ranked, each on all 100
tasks (Table~\ref{tab:variants}).

\textbf{Repeated creation.} GPT-6 Astra, GPT-5.6 Sol, and DeepSeek V4 Flash each
created all 100 tasks three times. Their 100-task means move by at most about
two points across runs, while a single task moves more: net of the judge's own
scoring noise (Appendix~\ref{app:repeat}), its score varies across runs with a
standard deviation of $3.7$, $5.0$, and $9.8$. Resampling tasks and runs
together, GPT-6 Astra leads GPT-5.6 Sol by $5.8$ $[4.2, 7.4]$ and GPT-5.6 Sol
leads DeepSeek V4 Flash by $12.1$ $[9.3, 14.9]$
(Appendix~\ref{app:repeatcreation}).

\textbf{Reasoning effort and direct answering.} GPT-6 Astra stays above every
other configuration at every effort level, and xhigh costs twice as much as
high for $0.9$ points; GLM-5.3 Flash keeps rising with effort, while DeepSeek V4
Flash saturates at high. In the direct-answer baseline the model writes the
palette and every pixel row in one response, with no tools and no view of the
result. GPT-6 Astra then stays close to its agent score on canvases of 32
pixels and below ($80.8$ against $81.4$) and falls behind above 64 pixels
($69.1$ against $83.1$), and DeepSeek V4 Flash completes none of the 14 tasks
of 128 pixels and above (Appendix~\ref{app:effort}).

\textbf{Image observation and cost.} Without image observation, GPT-6 Astra's
generation score moves from $82.3$ to $80.5$ but its editing score falls from
$88.0$ to $77.8$: an edit starts from an artwork someone else drew, and seeing
it lets the model understand it. GPT-6 Astra also uses the fewest output tokens
of any configuration, at \$0.24 per task (Appendix~\ref{app:cost}), and never
uses the full write budget.

\section{Limitations and Future Work}

Every configuration in this paper creates in one environment, whose prompt
carries no pixel-art technique guidance and no exemplar artwork. That
minimality is what makes the leaderboard a comparison between models rather
than between prompt designs, and it leaves the method itself unexplored: we do
not vary the tools, the budget, or what the model is told about the medium.
Those are the variables an agent designer would tune, and the environment we
release is where they can be tuned. PixelBench measures pixel-art creation
under one programmatic protocol; other environments and agent frameworks are
left to future work. Teaching a model the craft of the
medium, giving it references or critique to work against, or letting it revise
for longer are all directions this benchmark can now measure, on the way to
pixel art that enters a production workflow as it is.

\section{Conclusion}

We propose PixelBench, which formalizes pixel-art creation for language models
as generation and editing under a specified canvas size and color ceiling,
provides an agentic creation environment, and establishes an evaluation protocol
grounded in complete human rankings of 1,200 artworks by ten raters.
Under leave-one-out, the calibrated judge's agreement with the
consensus order falls within the range the ten raters span, overall and on
every slice, or above it, which lets it carry continued comparison on this
benchmark.

\subsubsection*{Reproducibility Statement}

We will release the 100 tasks (Appendix~\ref{app:tasks}), all 1,200 artworks,
the complete creation transcript of every task with the tokens, time, and
cost of every model call, the 1,000 human ranking lists, all judge responses,
the repeat scores of Appendix~\ref{app:repeat}, the artworks and scores of
the additional runs of Section~\ref{sec:variants}, the calibrated judge protocol with its full prompts
(Appendix~\ref{app:prompts}), and the creation environment, its sandbox,
and agent skill (Section~\ref{sec:environment}). Model APIs are not deterministic and human
rankings cannot be rerun identically, so a new run will differ in detail; the
released artworks, rankings, and judge responses let the results in every table
and figure be recomputed exactly.

\subsubsection*{AI Use Statement}

In this work, we used generative AI tools for generating synthetic data
(adding descriptive detail to the hand-written task statements, and rendering
the reference images of Section~\ref{sec:human}) and implementing methods (the
creation environment, the drawing API, and the evaluation scripts).
Additionally, we used generative AI tools to draft and edit parts of the paper,
write figure scripts, and identify related literature. We have reviewed all
AI-assisted work: the authors checked every task statement, verified every
citation against its source, and recomputed every reported number from the
archived results. During the revision, generative AI tools also
helped write the analysis scripts for the added experiments and draft the
revised text; every added number was recomputed from the archived results and
checked by the authors. We take responsibility for the final content of this work,
including text, claims, and artifacts produced with the aid of generative AI.

\bibliographystyle{iclr2027_conference}
\bibliography{references}

\appendix

\section{Full gallery}
\label{app:gallery}

Figures \ref{fig:gallery1} to \ref{fig:gallery6} show every artwork in the
benchmark: the 75
text-to-image tasks grouped by subject with characters and creatures first,
then the 25 editing tasks grouped into instruction editing, animation, and
multiview. Each row is one task. The leftmost column is the
reference: for a generation task the raw GPT-Image-2 rendering, and for an
editing task the unedited source artwork. The twelve columns to its right are
the twelve evaluated configurations in the order of Table~\ref{tab:main}, by
mean judge score over all 100 tasks from best to worst. Editing tasks show only the delivered result, with the frames of
a multi-frame result laid out left to right. Under each artwork, the first
number is its score from the calibrated judge and the second its mean rank over
the ten raters for that task (1 is best, 12 is worst). Both badges are filled
on the same red--yellow--green ramp, from worst to best, so a row can be read
at a glance. The GPT-Image-2 renderings are shown for reference only and
carry no score. Every artwork is drawn at its native resolution
as solid vector rectangles rather than as a bitmap, so the blocks stay exact at
any zoom level.

\input{figures/gallery_pages_b2}

\clearpage
\section{Tasks}
\label{app:tasks}

This appendix lists all 100 tasks in the order of the gallery in
Appendix~\ref{app:gallery}: the text-to-image tasks grouped by subject, then
the editing tasks grouped by type (instruction editing, animation, multiview).
Each text-to-image card gives the English task statement exactly as the model
receives it, the native canvas, and the color ceiling. Each
editing card adds the source artwork, enlarged with nearest-neighbor sampling,
the frames selected from it, and a citation of the work it comes
from. Editing tasks have no
color ceiling, and the model may enlarge the canvas; all frames of a
multi-frame result share one size.

\paragraph{Notice on source images.} The editing tasks start from existing
pixel art, and three of them from drawings by the authors. Each card below
cites the work its task uses: the platform it was published on, the author
where the source names one, and the exact files and frames taken. Each task
takes at most three frames of a single work, and the originals appear here, and
in the gallery of Appendix~\ref{app:gallery}, as citations: they document what
each task asks for and what the submissions were asked to change, and they are
reproduced for no other purpose. If you hold rights in one of these works and
would prefer it not to appear, please contact us: we will replace the affected
task with a similar new task and re-evaluate it.

\input{figures/tasks_appendix_b2}

\clearpage
\section{Judge prompts}
\label{app:prompts}

Every judge call is a single user message: a text block, followed by image
slots, each preceded by a short caption. The text below is sent verbatim;
braces mark fields filled from the task, and square brackets mark text that
appears only under the stated condition. The \{prompt\} field is the task
statement exactly as Appendix~\ref{app:tasks} gives it, the same text the
model creating the artwork received. All four protocol versions of
Section~\ref{sec:calib} share this text and differ only in what they switch
on: the presentation step adds the transparent margin to every pixel-art
image, the rubric step inserts the rubric before the scoring instruction, and
the reference step adds the reference slot.

\paragraph{Development subset.} The wording above was settled on 23 of the 75
generation tasks, using an earlier set of artworks and rankings by members
of the research team, before any artwork of the evaluation was created. The rubric was
written against four of them, one per subject---\texttt{magnifying-glass},
\texttt{glow-jellyfish}, \texttt{e-scooter-rider} and
\texttt{subway-platform}---and then fixed and confirmed on
\texttt{character-emotion-set} and \texttt{sunken-shipwreck}. The other
seventeen appear in short probes of the score range, of the reference slot and
of the judge's reasoning effort: \texttt{ancient-dragon-battle},
\texttt{ancient-treant-boss}, \texttt{battery-status-set},
\texttt{cyberpunk-alley}, \texttt{dim-sum-chef}, \texttt{dragon-hatchling},
\texttt{east-hero-full-body}, \texttt{farm-sheep},
\texttt{female-lab-researcher-chibi}, \texttt{mad-cyborg-scientist},
\texttt{magic-forest-cottage}, \texttt{portable-cassette-player},
\texttt{retro-film-camera}, \texttt{sky-pirate-girl},
\texttt{tavern-interior}, \texttt{teleport-portal} and
\texttt{two-character-standoff}. No editing task took part in any of them.
On the 77 tasks outside the development subset, the final protocol agrees
with the ten raters at $0.552$, close to its $0.557$ over all 100
tasks.

\paragraph{Image preparation.} The frames of a multi-frame result are packed
into the most nearly square grid; a material tile is repeated $2\times2$; the
presentation step then adds a transparent margin of $1/16$ of the width and
height on the left and top; finally the image is scaled with nearest-neighbor
sampling so that its longer side is 512 pixels, preserving the aspect ratio.
The GPT-Image-2 reference, which is not on a pixel grid, is scaled with
Lanczos resampling instead.

\paragraph{Generation.} The text block is

\begin{promptbox}
Task brief: \{prompt\}\\
Canvas: \{size\}; max colors: \{max\_colors\}\\[0.5ex]
[materials:] The image repeats the same native tile in a 2x2 grid, without gaps
or flips, so that tile boundaries are visible. Regular repetition is not a
defect.\\[0.5ex]
[rubric step on:] \{subject rubric\} [characters, creatures, objects, scenes:]
\{size sentence\}\\[0.5ex]
Rate this single pixel art image from 0 to 100 for overall quality for the
task. Reply with only one number from 0 to 100.
\end{promptbox}

\noindent followed by the image slots

\begin{promptbox}
[reference step on:] Image 1, reference image (calibration aid, not the work
being scored). The next image is a reference for this task, generated by
GPT-Image-2, a diffusion image model. It is not true pixel art: it does not lie
on the pixel grid, and it uses its own resolution and proportions. Treat it
only as one indication of what a visually strong result for this task can look
like. Do not score the candidate on how closely it resembles the reference: the
candidate's overall style and design may legitimately differ from it. Do not
treat the reference as ground truth, and do not deliberately reward a drawing
style that resembles it. \{image\}\\[0.5ex]
Candidate image (the work being scored): \{image\}
\end{promptbox}

\paragraph{Subject rubrics.}
\begin{promptbox}
Characters: Consider natural pose, an expressive silhouette and coherent
anatomy. Favor harmonious colors, balanced shapes and a cohesive composition.
The figure should read as a specific character: facial features clear and well
proportioned, the hairstyle plausible, and the whole legible rather than
cluttered.\\[0.5ex]
Creatures: Consider an appealing subject, expressive features, and a plausible
design of form and movement. Favor harmonious colors, balanced shapes and a
cohesive composition.\\[0.5ex]
Objects: Prioritize whether the object is clear and recognizable: as pixel art
it should be identifiable at a glance. Consider a recognizable silhouette,
clear parts and balanced placement; shading is optional. Credit intentional
simplicity; extra colors, shading or detail help only when they clarify
form.\\[0.5ex]
Scenes: Prioritize whether the scene as a whole is clear and recognizable:
spatial depth, coherent perspective, lighting, separation of foreground from
background, and the visual appeal and decorative detail of the image. Credit
intentional simplicity; extra colors, shading or detail help only when they
clarify form.\\[0.5ex]
Icons and interface: Prioritize instant recognition at native size, a clear
and conventional metaphor, the visual appeal of the icon, and clean separation
from the background. When the task asks for a set, consistency of style,
alignment, visual weight and sizing across the set matters as much as any
single element.\\[0.5ex]
Materials: Judge material identity, continuity across the tile boundaries, and
the absence of abrupt seams or value breaks. The material should look natural
and should not draw undue attention to itself, while still being attractive
and clear.\\[0.5ex]
Effects: Prioritize the overall shape of the effect, a readable sense of
motion, energy or impact, a clear focal point, and shapes that express the
phenomenon rather than a static decoration. Color ramps and contrast should
support the effect's intensity.\\[0.5ex]
Size sentence, longer side $\leq 32$: At this size, every pixel should be
placed deliberately, so that it supports a clear and appealing form.\\
Size sentence, longer side $> 32$: Details should enrich the image while
preserving clear forms.
\end{promptbox}

\paragraph{Editing, animation, and multiview.} The text block is

\begin{promptbox}
Editing instruction: \{prompt\}\\
Original canvas: \{size\}\\[0.5ex]
The result may use a different canvas size or different margins from the
original; that by itself is not a defect. [multi-frame:] \{frame note\}\\[0.5ex]
[rubric step on:] Weigh three things together.\\
(1) Instruction compliance: does the result actually make the change the
instruction asks for, on the right subject and to the right extent?\\
(2) Preservation: do the character, objects, scene, style and palette that the
instruction does not ask to change stay consistent with the original?\\
(3) Craft: is the result clean and visually appealing pixel art?\\[0.5ex]
[animation:] This is an action sequence, so also weigh:\\
(4) Inter-frame continuity: do consecutive frames read as one continuous
motion, holding the subject's identity, scale and placement steady between
frames?\\
(5) Pose plausibility: is the pose in each frame physically plausible for the
described action, is the subject's movement and posture sound, and does the
sequence express that action clearly?\\[0.5ex]
[multiview:] This is a multi-view turnaround, so also weigh:\\
(4) Cross-view identity: do all views show the same character or object as the
original, holding proportions, palette, costume and distinctive features
consistent across views, and is each view a plausible projection of that same
subject?\\[0.5ex]
The first image is the original artwork. The last image is the result the
model produced for this instruction. Rate the result from 0 to 100 for overall
quality on this editing task. Reply with only one number from 0 to 100.
\end{promptbox}

\noindent followed by the image slots

\begin{promptbox}
Image 1, the original artwork: \{image\}\\[0.5ex]
[reference step on:] Image 2, reference image (calibration aid, not the work
being scored). \{reference note\} \{image\}\\[0.5ex]
Image \{n\}, the model's result (the work being scored): \{image\}
\end{promptbox}

\noindent The frame note states the layout, for example ``The result is a
3-frame sequence shown as a single image, with the frames in order left to
right.'' When the task asks the first frame to reproduce the original, that
frame is already image 1 and is not shown again; the note then reads ``This
task asks for \{total\} frames, of which the first reproduces the original.
Only the remaining generated frame is shown.'' The reference note for a
reference taken from the source asset is

\begin{promptbox}
The next image is a reference result for this task. It is one valid way to
satisfy the instruction, not a target to copy, and a result that differs from
it can be equally good or better. Use it only to gauge what a reasonable
outcome for this instruction looks like. Do not score the result on how closely
it resembles the reference.
\end{promptbox}

\noindent and for a GPT-Image-2 edit it is

\begin{promptbox}
The next image is a reference for this task, generated by GPT-Image-2, a
diffusion image model. It is not true pixel art: it changes the proportions and
resolution of the original, and it does not lie on the pixel grid. Treat it
only as a rough hint of an attractive visual outcome, not as a target to copy;
a result that differs from it can be equally good or better. Do not score the
result on how closely it resembles the reference: the result's overall style
and design may legitimately differ from it. Do not treat the reference as
ground truth, and do not deliberately reward a drawing style that resembles
it.
\end{promptbox}

\section{Repeat scoring}
\label{app:repeat}

The repeat experiment of Section~\ref{sec:judge} draws 12 of the 100 tasks
under a fixed seed, before any scoring: one per generation subject, two
instruction edits, two animations, and one multiview. All 12 submissions of
each task are kept, for 144 artworks, and each is scored three further times
under the final protocol---\texttt{gpt-6-astra}, reasoning effort xhigh, a
fresh context per call, request order shuffled within each pass. The prompt text and the image hashes of all three passes match those of
the original run byte for byte. All 432 calls returned a valid score.

The mean score of the 144 artworks is $74.3$, $74.7$, and $74.8$ in the
three passes, and their three-pass mean differs from the single-pass score of
the same artworks by $+0.47$ points on average. Per artwork, the standard
deviation of the three scores averages $1.94$ points and has a median of
$1.53$; $62.5\%$ of repeat pairs agree within 2 points and $87.7\%$ within 5.
Where two artworks are 10 or more points apart, repeat scoring keeps their
order $99.7\%$ of the time. Across the 12 tasks, the per-artwork standard
deviation averages between $0.91$ and $3.37$ points, and the orders two passes
give a task's 12 submissions agree at a mean Kendall $\tau_b$ of $0.748$
$[0.658, 0.817]$.

Whether two artworks keep their order under repeat scoring depends on how far
apart they are (Table~\ref{tab:repeat-gap}). Over all pairs of submissions
within a task and all pairs of
passes, the order strictly reverses in $9.7\%$ of comparisons; restricted to
pairs at least 5 points apart in the earlier pass the rate is $3.4\%$, and at
least 10 points apart, $0.3\%$.

\begin{table}[htb]
\caption{Order stability against the score gap. ``Reversed'' is the share of
comparisons in which the later pass strictly reverses the earlier one;
``tie changed'' the share in which a tie becomes a strict order or the
reverse.}
\label{tab:repeat-gap}
\centering
\small
\begin{tabular}{@{}rccc@{}}
\toprule
Gap in the earlier pass & Comparisons & Reversed & Tie changed \\
\midrule
$\geq 0$  & 2376 &  9.7\% &  8.9\% \\
$\geq 3$  & 1827 &  5.3\% &  3.0\% \\
$\geq 5$  & 1509 &  3.4\% &  1.8\% \\
$\geq 10$ &  970 &  0.3\% &  0.7\% \\
\bottomrule
\end{tabular}
\end{table}

To carry this to the full task set, we resample the per-artwork residuals of
the three passes within each subject and type, in the proportions of the 100
tasks---75 generation tasks over seven subjects, 13 instruction edits, 10
animations, 2 multiviews---and rebuild each configuration's 100-task mean
$200{,}000$ times. This isolates the judge's sampling: the artworks and the
task set are held fixed. Table~\ref{tab:repeat-se} gives the resulting standard
error per configuration and the $95\%$ half-width of the difference between two
independent evaluations. Under the same resampling, 62 of the 66 pairs of
configurations keep the order of Table~\ref{tab:main} in at least $99\%$ of
draws; the four that do not are separated by at most $0.72$ points, and GPT-6
Astra's margin over every other configuration, at least $5.27$ points, holds in every draw. One pass per
artwork is therefore the protocol we release, and a gap below one point is
where a second pass earns its cost.

\begin{table}[htb]
\caption{Scoring standard error of a 100-task mean, by configuration, and the
$95\%$ half-width of the difference between two independent evaluations of the
same configuration.}
\label{tab:repeat-se}
\centering
\small
\begin{tabular}{@{}lcc@{}}
\toprule
Configuration & SE of 100-task mean & $95\%$ half-width of a rerun difference \\
\midrule
GPT-6 Astra              & 0.15 & 0.42 \\
Claude Opus 5            & 0.23 & 0.64 \\
Qwen3.8 Max              & 0.23 & 0.63 \\
Gemini 3.8 Flash         & 0.16 & 0.43 \\
GPT-5.6 Sol              & 0.25 & 0.69 \\
DeepSeek V4.1 Flash      & 0.26 & 0.71 \\
GLM-5.3 Flash            & 0.26 & 0.72 \\
Kimi K3                  & 0.18 & 0.50 \\
Gemini 3.1 Pro Preview   & 0.19 & 0.53 \\
DeepSeek V4 Pro          & 0.20 & 0.55 \\
DeepSeek V4 Flash        & 0.21 & 0.58 \\
Qwen3.5-397B-A17B        & 0.27 & 0.73 \\
\bottomrule
\end{tabular}
\end{table}

\section{Text observation}
\label{app:textobs}

\texttt{view} is the only channel through which a model obtains an image
observation (Section~\ref{sec:environment}). Five of the evaluated models take
no image input: DeepSeek V4 Flash, DeepSeek V4.1 Flash, DeepSeek V4 Pro, GLM-5.3 Flash, and
Qwen3.5-397B-A17B. GPT-6 Astra without image observation switches
vision off in the same way, and apart from \texttt{view} it shares tools,
prompt, and budget with the full configuration, so the difference between them
measures what image observation contributes to creation quality
(Section~\ref{sec:variants}).

In all of these cases \texttt{view} returns no image and the model reads the
artwork as a character picture instead. \texttt{matrix\_text()} prints the
canvas, or a region of it, as a character grid: each line of text is one row of
pixels and each character one pixel, the same color is always written with the
same character (taken from \texttt{a}--\texttt{z}, \texttt{A}--\texttt{Z},
\texttt{0}--\texttt{9} by palette index), and transparent pixels are written as
a dot. A legend after the grid maps each character to its color name, hex
value, and palette index, and the model can also print the current palette to
obtain the mapping between colors and characters. \texttt{stats\_text()} gives
the color count, coverage, bounding box, and most frequent colors. On large
canvases with many colors, the model can print a single region, or shrink a
selection before printing it.

\section{Other judge models}
\label{app:crossjudge}

The judge of Section~\ref{sec:judge} is GPT-6 Astra, which also leads both
leaderboards. To see what the protocol returns when a different model answers,
we score the artworks of the evaluation again with two other judge models:
Claude Opus 5, at reasoning effort low, and the open-weight DeepSeek V4.1
Flash, at high; the Astra judge runs at \texttt{xhigh}. Every call carries the
same prompt text and the same image hashes as the corresponding call of the
Astra judge, so the judges receive identical inputs. Claude Opus 5 returned no
score for two artworks of \texttt{fist-swing}, so its comparisons cover the
other 99 tasks. Table~\ref{tab:crossjudge} gives the results.

\begin{table}[htb]
\caption{Three judge models under the same protocol. Agreement is the
per-task mean Kendall $\tau_b$ with the ten-rater consensus order; leave-one-out
is the judge's mean agreement with the consensus of nine raters, with the
number of held-out raters below it in parentheses; $\rho$ is the Spearman
correlation of the judge leaderboard with the human one. The lower block gives
the per-task paired difference in agreement, GPT-6 Astra minus the other judge,
with 95\% intervals. Claude Opus 5's values cover the 99 tasks it scored in
full, 24 of them editing.}
\label{tab:crossjudge}
\centering
\footnotesize
\setlength{\tabcolsep}{4pt}
\begin{tabular}{@{}llccccc@{}}
\toprule
Judge & Reasoning effort & All & Generation & Editing & Leave-one-out & $\rho$ \\
\midrule
GPT-6 Astra         & xhigh & 0.557 & 0.563 & 0.538 & 0.554 (3) & 0.958 \\
Claude Opus 5       & low   & 0.498 & 0.469 & 0.590 & 0.494 (1) & 0.825 \\
DeepSeek V4.1 Flash & high  & 0.414 & 0.407 & 0.432 & 0.413 (0) & 0.958 \\
\bottomrule
\end{tabular}

\vspace{6pt}
\begin{tabular}{@{}lccc@{}}
\toprule
GPT-6 Astra minus   & All & Generation & Editing \\
\midrule
Claude Opus 5       & $+0.058$ $[+0.016, +0.098]$ & $+0.094$ $[+0.050, +0.136]$ & $-0.056$ $[-0.146, +0.037]$ \\
DeepSeek V4.1 Flash & $+0.144$ $[+0.108, +0.179]$ & $+0.156$ $[+0.115, +0.197]$ & $+0.106$ $[+0.041, +0.170]$ \\
\bottomrule
\end{tabular}
\end{table}

\textbf{Agreement within a task.} GPT-6 Astra agrees with the consensus order
more than either other judge, and under leave-one-out it is above three of the
ten raters, against one for Claude Opus 5 and none for DeepSeek V4.1 Flash.

\textbf{Leaderboards.} Both judges put GPT-6 Astra first, and their
leaderboards correlate with the Astra judge's at Spearman $\rho = 0.895$
(Claude) and $0.972$ (DeepSeek). The DeepSeek judge puts Claude Opus 5 second
and Qwen3.8 Max third, and places the three DeepSeek configurations seventh,
tenth, and eleventh, exactly where the raters place them; its leaderboard
correlates with the human one as closely as the Astra judge's, so it
reproduces the leaderboard more closely than the order within each task. The
Claude judge puts Claude Opus 5 second, as the Astra judge does, and departs
from the raters most on Gemini 3.8 Flash, which it places eighth and the
raters second. Their mean scores sit about 17 (Claude) and 14 (DeepSeek)
points below the Astra judge's, so the scales are not interchangeable and the
order is the comparable quantity.

\section{Slices by subject, canvas size, and editing type}
\label{app:slices}

Table~\ref{tab:slices} gives the full breakdown behind
Section~\ref{sec:slices}: the 75 generation tasks by subject and by canvas
size, with the four quantities that section reads off them, and the 25
editing tasks by type. The last three columns repeat the leave-one-out
comparison of Section~\ref{sec:calib} within each slice: each of the ten
raters, and the judge, against the mean rank of the other nine. The judge
falls within the range of the ten raters on every slice, and above all ten on
canvases larger than 64 pixels. Canvas sizes appear in two groupings, the four
bands of Section~\ref{sec:slices} and the two bands of 16 to 32 and of 64
pixels and above. Each slice is scored on the tasks it contains, and the task
counts of Section~\ref{sec:bench} apply.

\begin{table}[htb]
\caption{Tasks by subject, canvas size (longer side), and editing type.
Score SD is the standard deviation of the judge scores of a task's twelve
artworks, averaged over the slice; rater--rater is the pairwise agreement among
the ten raters; rank SD is the standard deviation of the twelve
configurations' mean human ranks over the slice; judge--human is the
judge against the ten-rater consensus. Leave-one-out gives the judge's mean
agreement with the consensus of nine raters, the range of the ten held-out
raters, and how many of them fall below the judge.}
\label{tab:slices}
\centering
\footnotesize
\setlength{\tabcolsep}{3.5pt}
\input{tables/table_slices_b2}
\end{table}

\section{Evaluation runs}
\label{app:runs}

All twelve configurations of Table~\ref{tab:main} created all 100 tasks
under the environment of Section~\ref{sec:environment}, with a budget of 25
source writes for generation and 15 for editing, and the same system prompt,
drawing API, and tools. Every model call was logged with a digest of the
request, its token usage (input, cached input, and output including
reasoning), its latency, and its retries; the final
context and every image sent to the model are stored with the transcript.

\textbf{Reruns.} A failure of the serving infrastructure (a 5xx or 429
response, an exhausted quota, a dropped connection) says nothing about the
model, and a task that hit one was run again from the start until it
completed. When the model itself returned no artwork, was cut off (a single
request longer than 75 minutes or a task longer than three hours), or submitted
fully transparent frames, the task was rerun once with the same
configuration, and the rerun counts. Of the 1,200 artworks, 1,190 were
submitted formally by the model; for the other ten, all of which used the full
write budget without a formal submission, the environment recovered the
artwork from the canvases the program had drawn. No artwork is empty.

\textbf{Sandbox.} The sandbox of Section~\ref{sec:environment} confines a
program to its own run directory and blocks network access and the start of
other programs. It ships with the environment.

\section{Human rating}
\label{app:raters}

\begin{figure}[htb]
\centering
\includegraphics[width=0.49\linewidth]{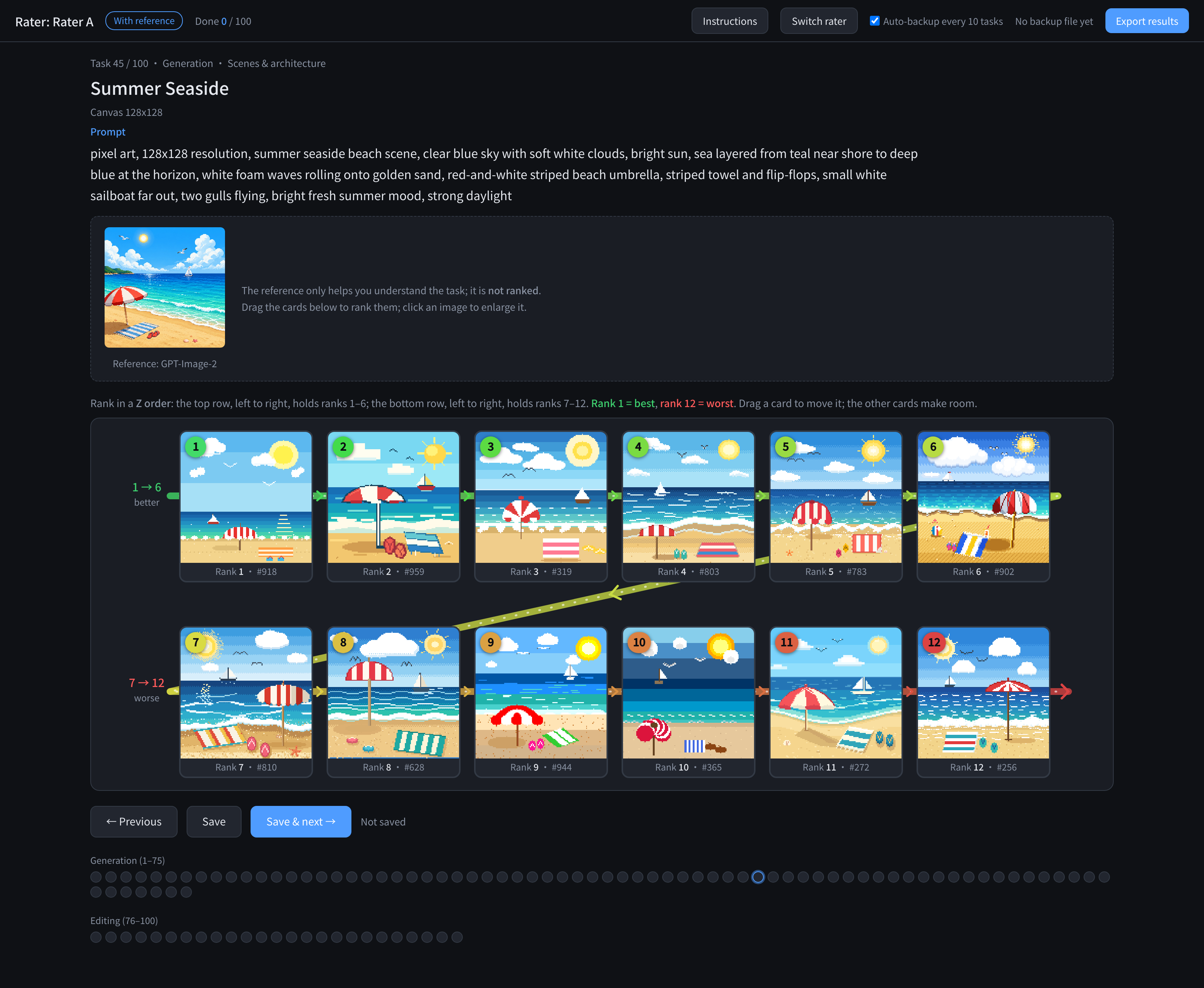}\hfill
\includegraphics[width=0.49\linewidth]{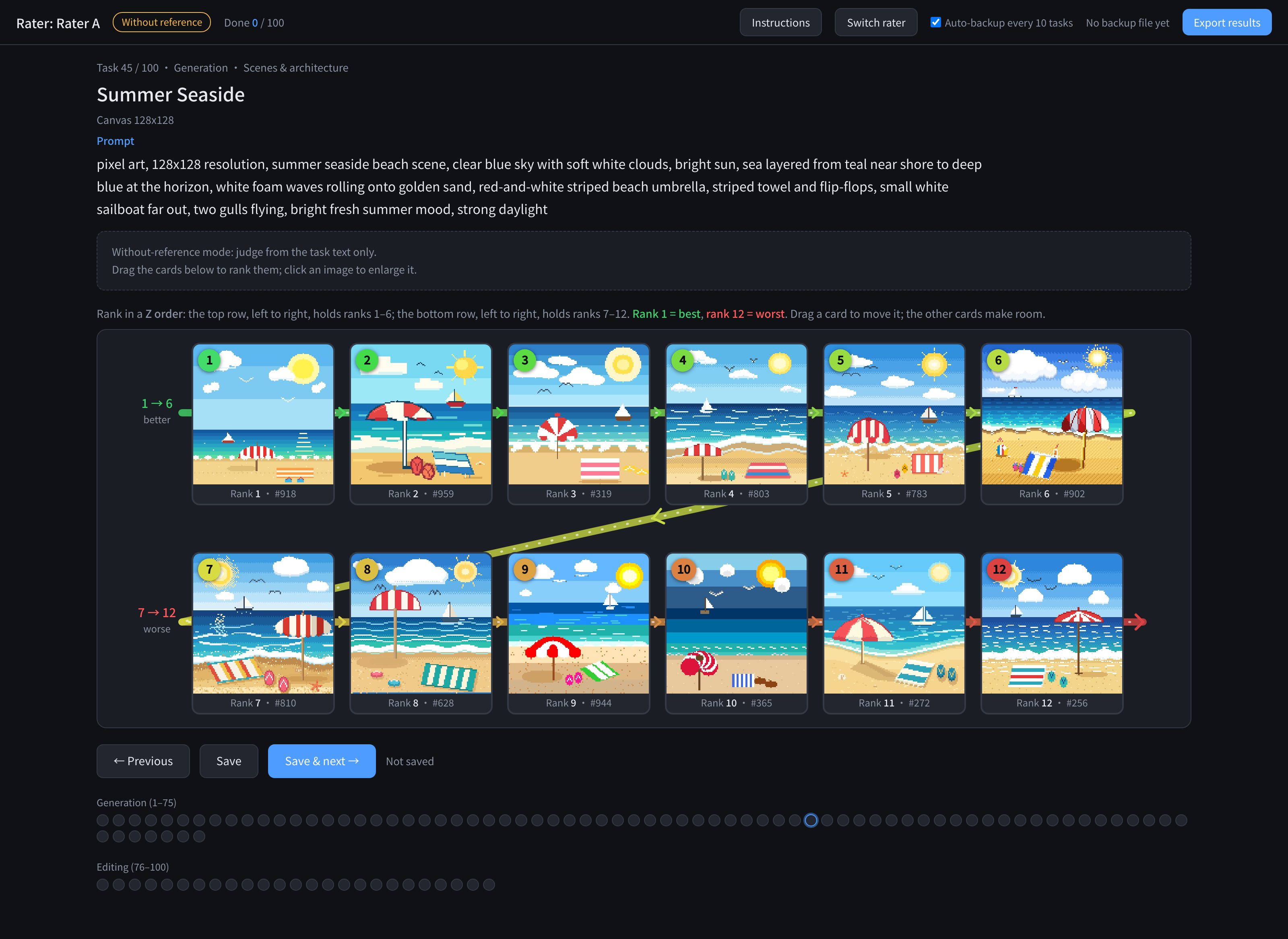}
\caption{The ranking interface on \texttt{summer-seaside}, with the
reference image (left) and without it (right). Cards carry random labels; the
rater drags them into a Z-shaped order from best (top left) to worst (bottom
right).}
\label{fig:rating-ui}
\end{figure}

\textbf{Raters and interface.} The ten raters of Section~\ref{sec:human}
all have professional art backgrounds, mainly in art and game design, with
experience in graphic design, years of painting, or pixel-style game art. Each
ranked all 100 tasks in an offline browser page (Figure~\ref{fig:rating-ui}).
For each task the page shows the task statement and canvas requirement, the
reference images of the rater's mode, and the twelve artworks as cards under
random labels, magnified with nearest-neighbor sampling, with animations
playing in a loop. The rater drags the cards into a Z-shaped order, ranks 1 to
6 in the top row and 7 to 12 in the bottom row, and saves. The initial order of
the cards is drawn at random for every rater and task, and the page records it
together with the time spent on the task.

\textbf{Position and time.} The initial order has little effect on the final
ranking: the Kendall $\tau_b$ between where a card started and where it was
ranked averages $0.10$ over raters, between $0.02$ and $0.21$ for individual
raters. Raters spent a median of about 70 seconds on a task. The two raters
who ranked without a reference reach leave-one-out agreements of $0.620$ and
$0.611$, fourth and fifth of the ten.

\textbf{Without a reference.} Two of the ten raters ranked all 100 tasks
without any reference: for generation they saw no GPT-Image-2 rendering, and
for editing only the original. Taking the mean ranks of the eight raters with
references and of the two without as two separate leaderboards, the two
correlate at Spearman $\rho = 0.972$ over all 100 tasks and $0.965$ over the 75
generation tasks, with GPT-6 Astra first on both. On generation a
configuration's mean rank shifts by $0.35$ places on average. Among the 45 ways
of splitting the ten raters into two and eight, the other 44 give generation
leaderboards that correlate at $0.930$ to $0.993$, and the split by reference
falls inside this range, so the human leaderboard barely depends on the
reference. On the judge side, the protocol versions before
and after the reference step agree with the two groups as follows: without the
reference the judge reaches $0.536$ against the eight and $0.504$ against the
two, and with it $0.549$ and $0.514$, a similar change for both groups.

\section{Analyses scored by the judge}
\label{app:variants}

This appendix gives the details behind Section~\ref{sec:variants}. Every
run covers all 100 tasks and is scored once by the judge.

\subsection{Repeated creation}
\label{app:repeatcreation}

Every configuration in Table~\ref{tab:main} creates each task once. To
measure the randomness of creation itself, three configurations created all
100 tasks two more times, from scratch and under the same settings: GPT-6
Astra, which leads the table; GPT-5.6 Sol, from its middle; and DeepSeek V4
Flash, an open-weight model. The judge
scored every artwork once. Table~\ref{tab:repeatcreation} gives the results.
The per-task standard deviation is taken across the three runs and averaged
over tasks as a variance; removing the judge's own repeat-scoring variance
(pooled standard deviation $2.30$, Appendix~\ref{app:repeat}) leaves the part
that comes from creation. The paired intervals resample tasks and, within each
resampled task, one run of each configuration, 10{,}000 times; they cover
both the sampling of tasks and the randomness of creation.
Figure~\ref{fig:gallery_variants}a shows the three runs side by side on a
random sample of tasks.

\begin{table}[htb]
\caption{Three independent creation runs. Run 1 is the run of
Table~\ref{tab:main}. Per-task SD is the standard deviation of a task's score
across runs; creation SD removes the judge's repeat-scoring variance from it;
$\rho$ is the range of the Spearman correlation of the 100 task scores between
two runs, over the three pairs of runs.}
\label{tab:repeatcreation}
\centering
\small
\setlength{\tabcolsep}{4pt}
\begin{tabular}{@{}lccccc@{}}
\toprule
Configuration & Runs 1 / 2 / 3 & SD of means & Per-task SD & Creation SD & $\rho$ \\
\midrule
GPT-6 Astra (high)       & 83.76 / 82.95 / 81.69 & 1.04 & 4.35 & 3.69 & 0.69--0.77 \\
GPT-5.6 Sol (medium)     & 76.92 / 77.38 / 76.72 & 0.34 & 5.46 & 4.95 & 0.67--0.78 \\
DeepSeek V4 Flash (high) & 65.52 / 63.95 / 65.37 & 0.87 & 10.02 & 9.76 & 0.54--0.62 \\
\bottomrule
\end{tabular}

\vspace{6pt}
\begin{tabular}{@{}lc@{}}
\toprule
Paired difference, resampling tasks and runs & Mean [95\% CI] \\
\midrule
GPT-6 Astra $-$ GPT-5.6 Sol           & $+5.79$ $[+4.24, +7.41]$ \\
GPT-5.6 Sol $-$ DeepSeek V4 Flash      & $+12.06$ $[+9.34, +14.87]$ \\
GPT-6 Astra $-$ DeepSeek V4 Flash      & $+17.85$ $[+14.92, +21.00]$ \\
\bottomrule
\end{tabular}
\end{table}

\subsection{Reasoning effort and direct answering}
\label{app:effort}

\begin{figure}[p]
\centering
\includegraphics[width=0.93\textwidth]{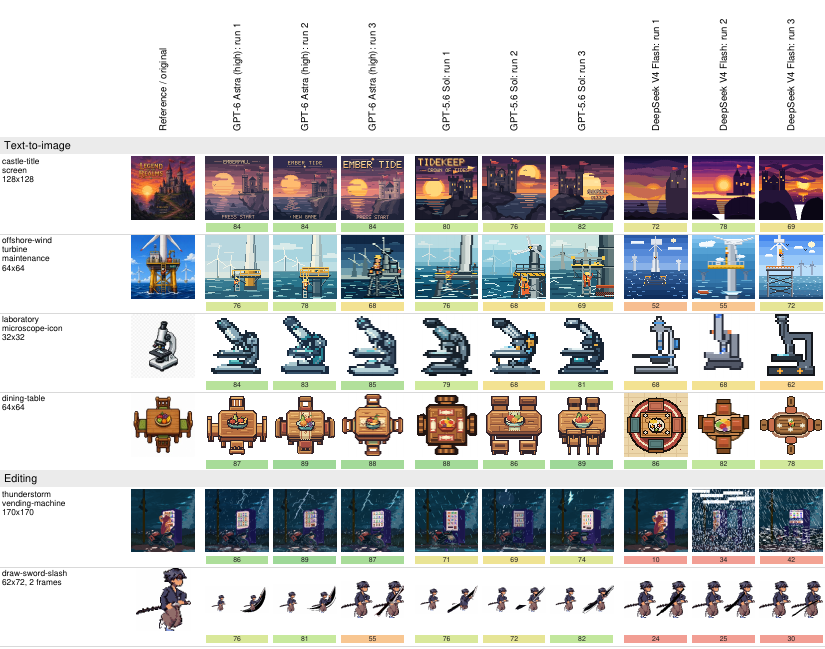}\\[2pt]
{\small (a) Repeated creation}\\[8pt]
\includegraphics[width=0.93\textwidth]{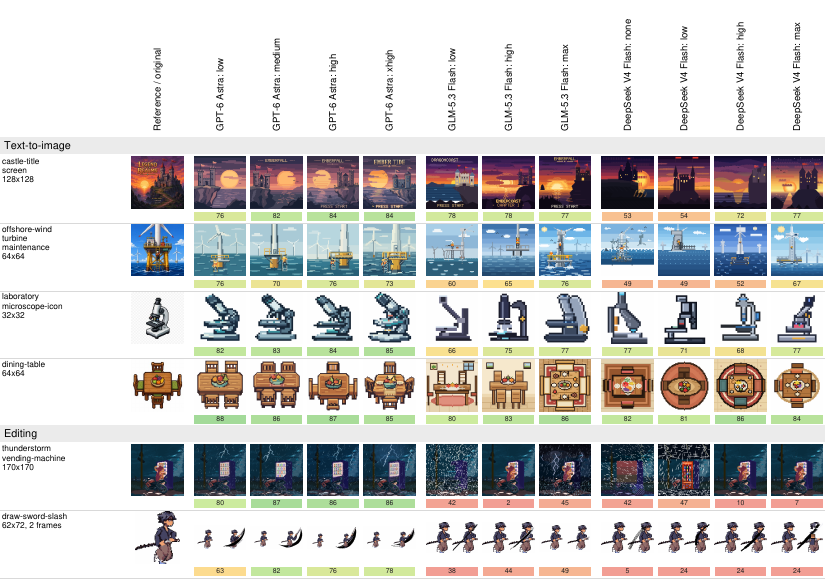}\\[2pt]
{\small (b) Reasoning effort}
\caption{Sample of the analyses scored by the judge, on six tasks drawn at
random (four text-to-image, two editing). (a) Each configuration created every
task three times from scratch under the same settings; run 1 is the run of
Table~\ref{tab:main}. (b) The reasoning-effort levels of three models; the
levels of Table~\ref{tab:main} are GPT-6 Astra at high, DeepSeek V4 Flash at
high, and GLM-5.3 Flash at max. Badges: judge score.}
\label{fig:gallery_variants}
\end{figure}

\textbf{Reasoning effort.} Three models were run on all 100 tasks at
further reasoning efforts, with everything else as in Table~\ref{tab:main}
(Table~\ref{tab:effort}): GPT-6 Astra at low, medium, and xhigh besides high;
DeepSeek V4 Flash with reasoning switched off and at low and max besides high;
and GLM-5.3 Flash at low and high besides max. These are the levels each model
accepts. Every level submitted formally on at least 94 of
the 100 tasks, so the differences come from the artworks rather than from
missing submissions. GPT-6 Astra rises slowly with effort and stays above
every other configuration of Table~\ref{tab:main} at every level; xhigh costs
about twice as much as high for $0.9$ points. GLM-5.3 Flash rises at every
step, reasoning about eleven times as much at max as at high. DeepSeek V4 Flash
at max reasons barely more than at high, a median of 33k against 30k tokens
per task, and scores the same within the interval.
Figure~\ref{fig:gallery_variants}b shows the levels side by side on the same
sample.

\begin{table}[htb]
\caption{Reasoning effort, each level on all 100 tasks. $\Delta$ is the
paired difference from the level of Table~\ref{tab:main} (marked $^\ast$) with
a 95\% task-bootstrap interval; formal counts tasks submitted by the model
itself; output is the median per task, reasoning included; cost is per task at
list price (Appendix~\ref{app:cost}).}
\label{tab:effort}
\centering
\small
\setlength{\tabcolsep}{4pt}
\begin{tabular}{@{}llccccccc@{}}
\toprule
Model & Level & All & Gen. & Edit & $\Delta$ [95\% CI] & Formal & Output (k) & Cost \\
\midrule
GPT-6 Astra       & low         & 81.8 & 80.5 & 85.8 & $-1.9$ $[-2.6, -1.3]$ & 100 & 1.3 & \$0.19 \\
                  & medium      & 82.4 & 80.8 & 87.4 & $-1.3$ $[-2.1, -0.6]$ & 100 & 1.3 & \$0.17 \\
                  & high$^\ast$ & 83.8 & 82.3 & 88.0 & ---                   & 100 & 1.9 & \$0.24 \\
                  & xhigh       & 84.7 & 83.6 & 87.9 & $+0.9$ $[+0.3, +1.5]$ & 100 & 4.7 & \$0.50 \\
\midrule
DeepSeek V4 Flash & none        & 62.4 & 64.9 & 54.8 & $-3.2$ $[-6.0, -0.3]$ &  96 & 19.8 & \$0.05 \\
                  & low         & 61.0 & 62.9 & 55.2 & $-4.5$ $[-7.8, -1.6]$ & 100 & 11.0 & \$0.03 \\
                  & high$^\ast$ & 65.5 & 66.9 & 61.4 & ---                   &  99 & 35.0 & \$0.06 \\
                  & max         & 66.1 & 68.0 & 60.5 & $+0.6$ $[-2.0, +3.2]$ &  97 & 37.4 & \$0.06 \\
\midrule
GLM-5.3 Flash     & low         & 62.6 & 65.1 & 55.1 & $-12.3$ $[-15.0, -9.9]$ & 96 & 2.1 & \$0.01 \\
                  & high        & 69.8 & 72.2 & 62.8 & $-5.1$ $[-7.8, -2.6]$  & 94 & 14.6 & \$0.02 \\
                  & max$^\ast$  & 75.0 & 75.4 & 73.6 & ---                    & 100 & 137.6 & \$0.12 \\
\bottomrule
\end{tabular}
\end{table}

\textbf{Direct answering.} The direct-answer baseline removes the
environment altogether. The model receives the task statement, and for an
editing task the original written out in the same text format, and answers
once: a palette block that assigns a one-character code to each color, then one
block per frame with one row of codes per pixel row at the native canvas size.
There are no tools, no execution, no view of the result, and no second turn.
An answer may use up to 131{,}072 tokens, reasoning included, and 60 minutes.
The parser pads or truncates a row that is off by at most two pixels or 5\% of
the width, and rejects a frame whose row count is off by more, more than a
fifth of whose rows are off by more, or more than 5\% of whose pixels use
undeclared codes; a task without a valid answer scores 0. The answers are
rendered and scored by the judge exactly like the agent's artworks.
Table~\ref{tab:direct} gives the results by canvas size, for GPT-6 Astra and
DeepSeek V4 Flash at the reasoning effort of Table~\ref{tab:main}.

\begin{table}[htb]
\caption{Direct answering against the agent, by canvas size (longer
side). Answered counts tasks with a valid answer; the direct-answer score
counts the others as 0.}
\label{tab:direct}
\centering
\small
\begin{tabular}{@{}lcccccc@{}}
\toprule
 & \multicolumn{3}{c}{GPT-6 Astra (high)} & \multicolumn{3}{c}{DeepSeek V4 Flash (high)} \\
\cmidrule(lr){2-4}\cmidrule(l){5-7}
Tasks & Answered & Direct & Agent & Answered & Direct & Agent \\
\midrule
Generation, $\leq 32$ (28)  & 28 & 80.8 & 81.4 & 25 & 43.2 & 70.9 \\
Generation, 33--64 (33)     & 32 & 76.0 & 82.8 & 13 &  2.2 & 65.2 \\
Generation, $> 64$ (14)     & 14 & 69.1 & 83.1 &  0 &  0.0 & 62.9 \\
\midrule
Generation (75)             & 74 & 76.5 & 82.3 & 38 & 17.1 & 66.9 \\
Editing (25)                & 24 & 78.1 & 88.0 & 14 & 23.0 & 61.4 \\
All (100)                   & 98 & 76.9 & 83.8 & 52 & 18.6 & 65.5 \\
\bottomrule
\end{tabular}
\end{table}

\subsection{Cost}
\label{app:cost}

\begin{figure}[htb]
\centering
\includegraphics[width=\linewidth]{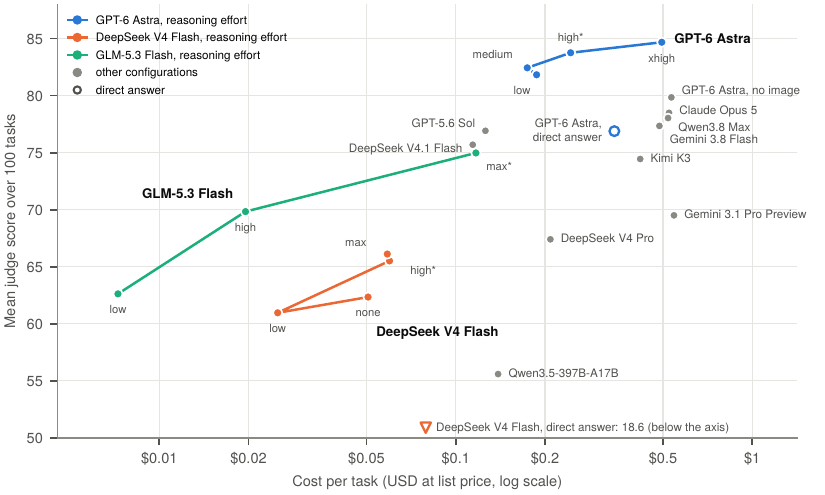}
\caption{Mean judge score over all 100 tasks against cost per task at list
price. Colored lines join the reasoning-effort levels of one model, with an
asterisk at the level of Table~\ref{tab:main}; grey points are the other
configurations at the settings of Table~\ref{tab:main}, including GPT-6 Astra
without image observation; hollow points are the direct-answer baseline of
Appendix~\ref{app:effort}.}
\label{fig:cost}
\end{figure}

Every model call of the evaluation was logged with its token usage, its
latency, and its retries. Table~\ref{tab:cost} averages them over the 100
tasks of each configuration. Cost is computed at the list price of each model in
US dollars per million tokens, retrieved on 2026-11-10, with cached input at
the cached rate. Minutes are median wall-clock time per task and depend mostly on the
provider's load at the time; running the programs takes a negligible share.
Figure~\ref{fig:cost} plots each configuration's score against its cost per
task, together with the reasoning-effort levels of Appendix~\ref{app:effort}.

\begin{table}[htb]
\caption{Per-task averages over the 100 tasks. Calls are model calls;
writes are source modifications; input and output are thousands of tokens,
output including reasoning; cached is the share of input served from the
provider's cache.}
\label{tab:cost}
\centering
\small
\setlength{\tabcolsep}{4pt}
\input{tables/table_cost_b2}
\end{table}

\section{Pixel art and SVG generation}
\label{app:svg}

\begin{figure}[p]
\centering
\begin{minipage}[t]{0.485\linewidth}\centering
\includegraphics[width=\linewidth,height=3.5cm,keepaspectratio]{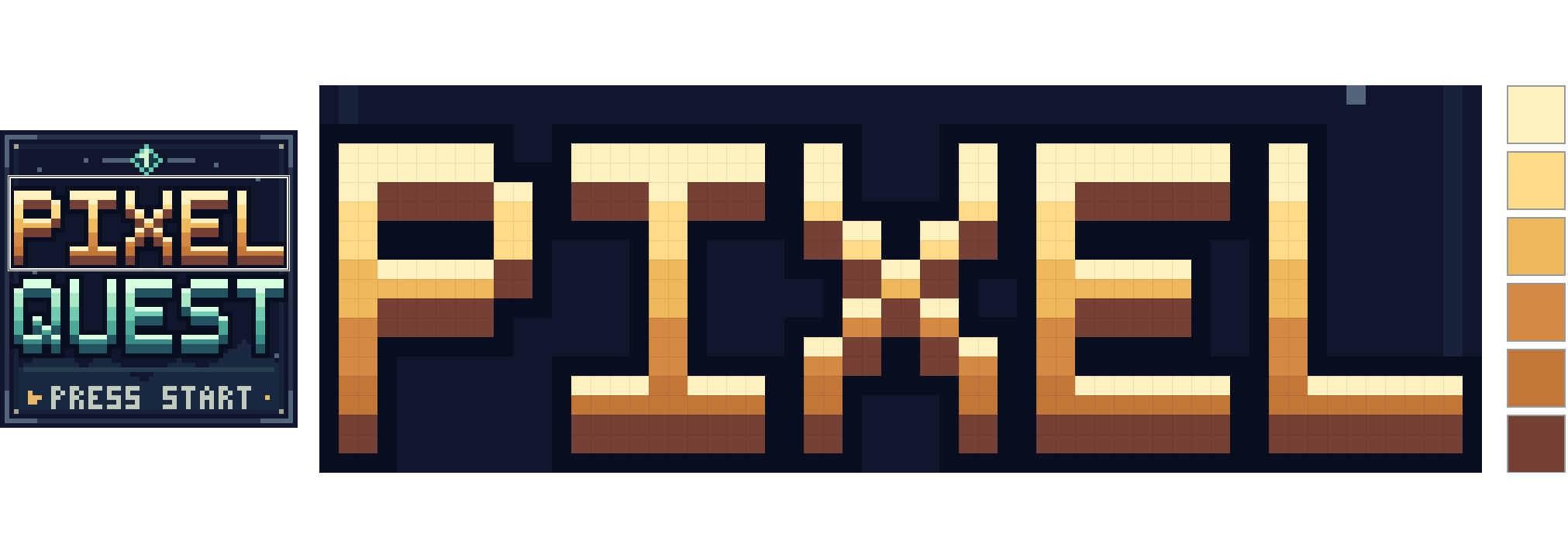}\\[2pt]
{\footnotesize\textbf{Banded gradient.} \texttt{typography}, GPT-6 Astra}
\end{minipage}\hfill
\begin{minipage}[t]{0.485\linewidth}\centering
\includegraphics[width=\linewidth,height=3.5cm,keepaspectratio]{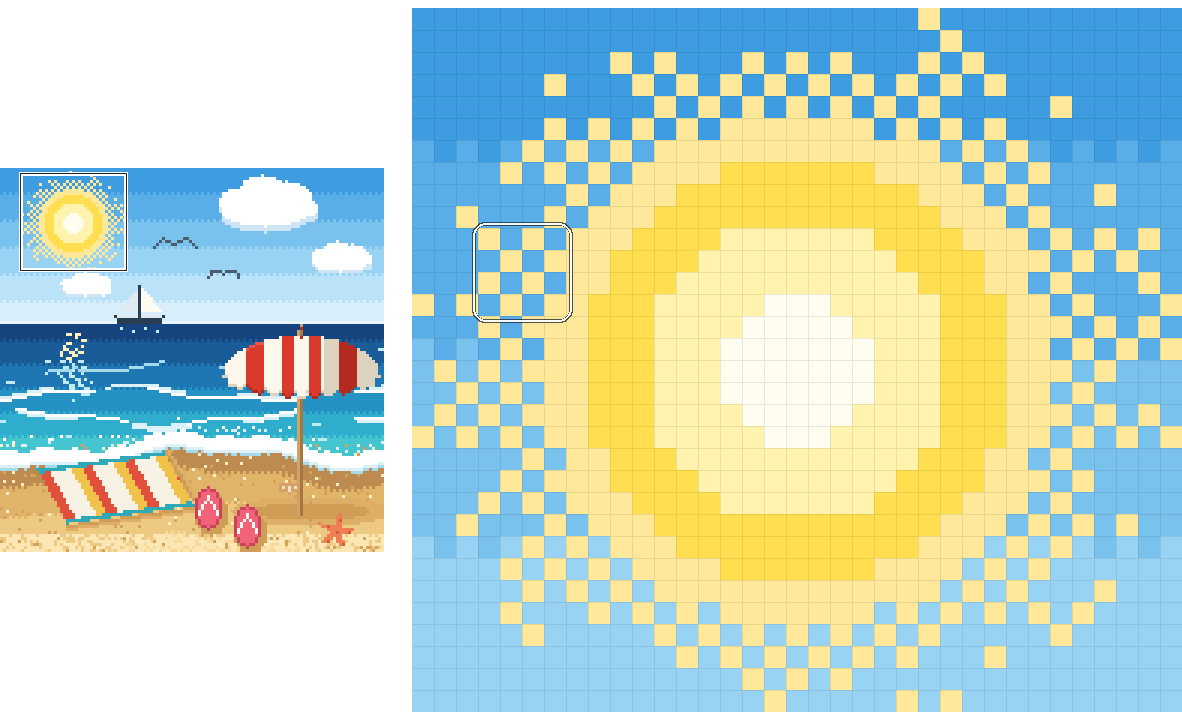}\\[2pt]
{\footnotesize\textbf{Dithering.} \texttt{summer-seaside}, Qwen3.8 Max}
\end{minipage}\\[8pt]
\begin{minipage}[t]{0.485\linewidth}\centering
\includegraphics[width=\linewidth,height=3.5cm,keepaspectratio]{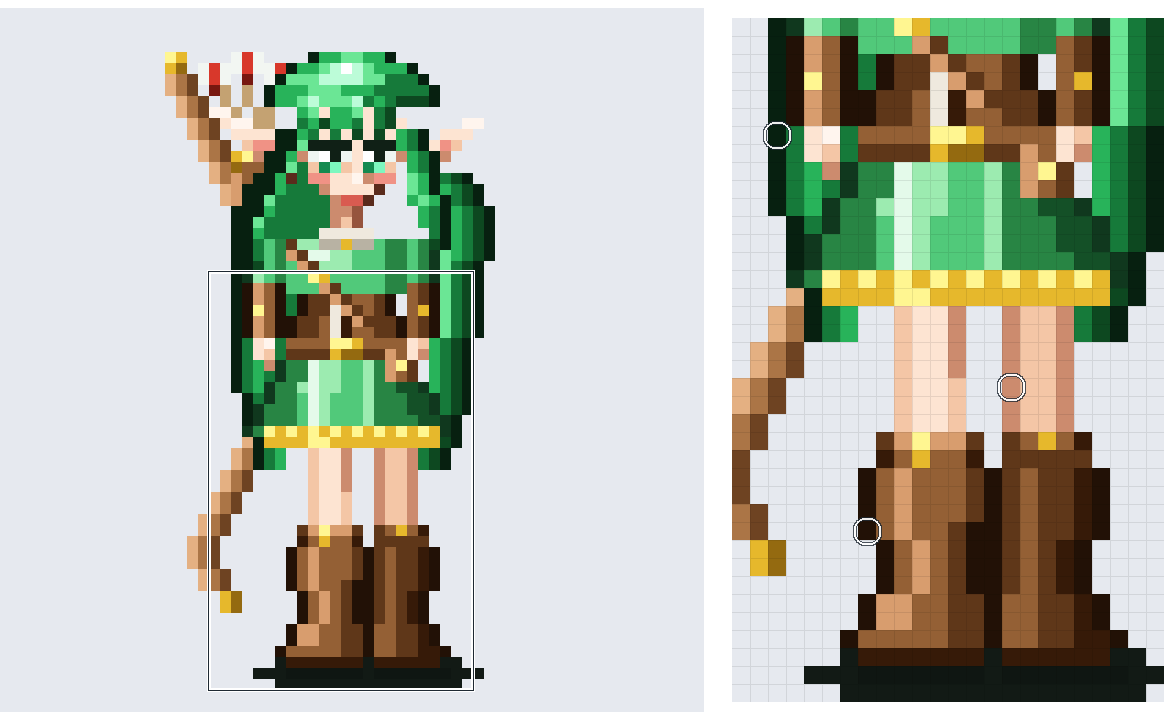}\\[2pt]
{\footnotesize\textbf{Selective outline.} \texttt{female-elf-archer}, Gemini 3.8 Flash}
\end{minipage}\hfill
\begin{minipage}[t]{0.485\linewidth}\centering
\includegraphics[width=\linewidth,height=3.5cm,keepaspectratio]{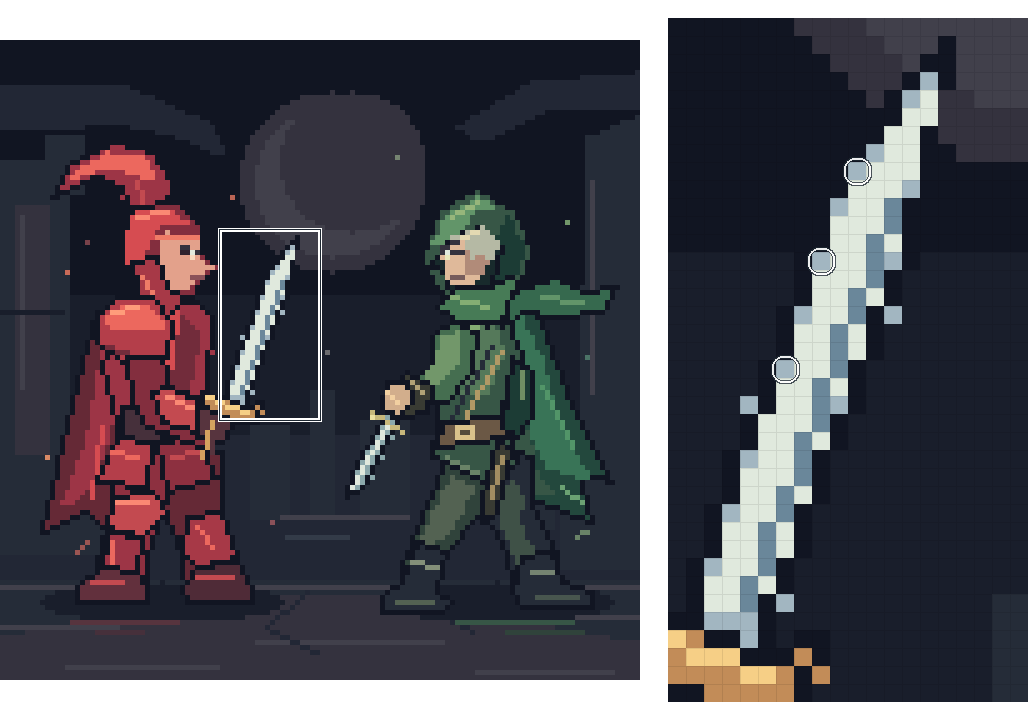}\\[2pt]
{\footnotesize\textbf{Manual anti-aliasing.} \texttt{two-character-standoff}, GPT-6 Astra}
\end{minipage}\\[8pt]
\begin{minipage}[t]{0.485\linewidth}\centering
\includegraphics[width=\linewidth,height=3.5cm,keepaspectratio]{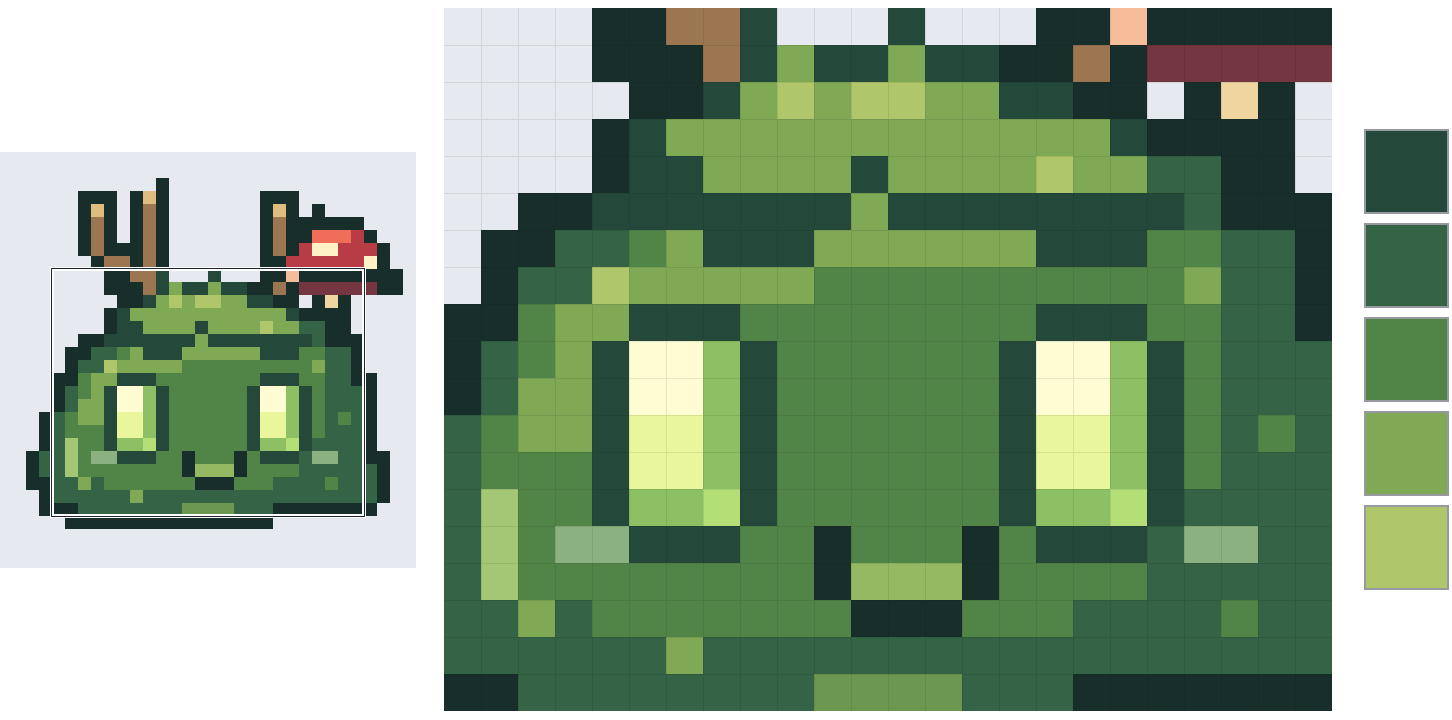}\\[2pt]
{\footnotesize\textbf{Hue-shifted ramp.} \texttt{forest-slime-monster}, GPT-6 Astra}
\end{minipage}\hfill
\begin{minipage}[t]{0.485\linewidth}\centering
\includegraphics[width=\linewidth,height=3.5cm,keepaspectratio]{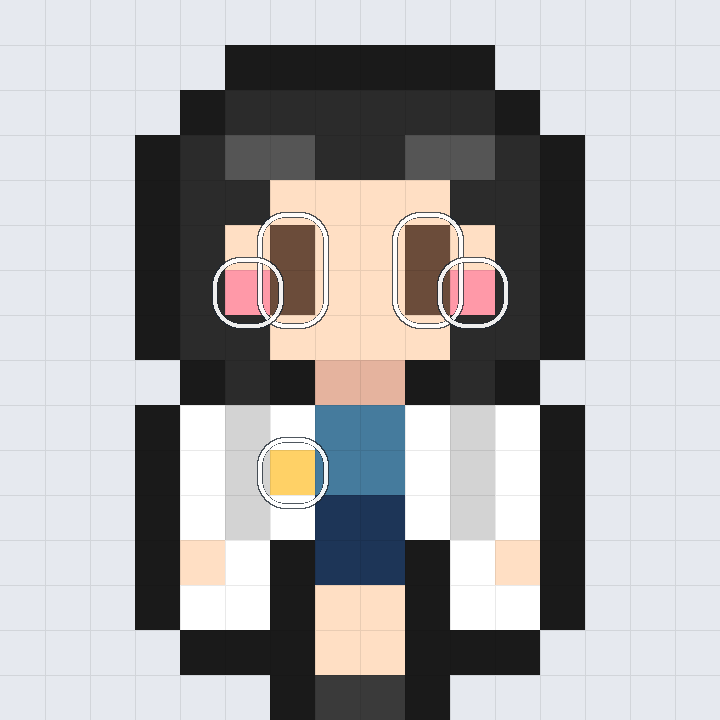}\\[2pt]
{\footnotesize\textbf{Single-pixel detail.} \texttt{female-lab-researcher-chibi}, Gemini 3.1 Pro Preview}
\end{minipage}\\[8pt]
\begin{minipage}[t]{0.485\linewidth}\centering
\includegraphics[width=\linewidth,height=3.5cm,keepaspectratio]{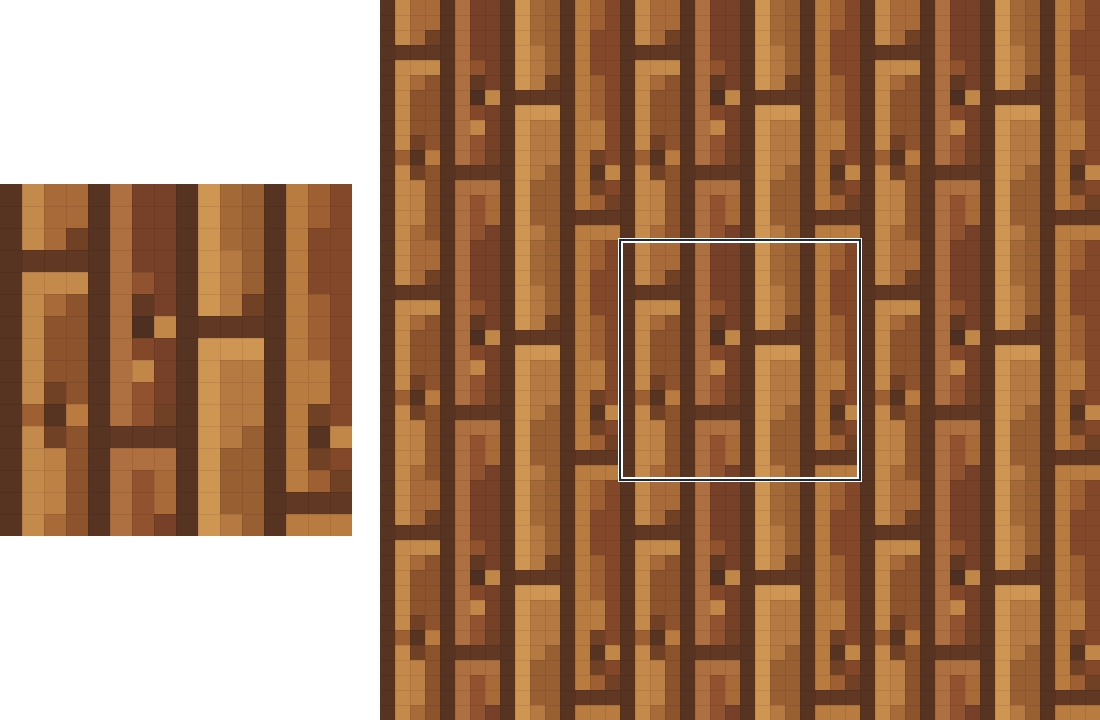}\\[2pt]
{\footnotesize\textbf{Seamless tiling.} \texttt{rpg-wooden-planks-material}, GPT-6 Astra}
\end{minipage}\\[8pt]
\caption{Pixel-art techniques in the submissions, each from a generation task
drawn from an empty canvas. In each panel the whole artwork is on the left,
with a frame around the region enlarged on its right; enlargements use
nearest-neighbor scaling with a faint pixel grid, and transparent pixels are
flat grey. Thin rings mark example pixels: a $4\times4$ checkerboard in the
halo of the sun; outline pixels whose color follows the region they bound,
dark green on the cloak, a darker skin tone on the leg, and dark brown on the
boot; intermediate tones at the steps of a diagonal blade; and eyes of
$1\times2$ pixels with a one-pixel blush and pen on a $16\times16$ canvas.
Swatches give the five bands and the shadow of the letters, and the slime's
body ramp from shadow to highlight, which shifts from blue-green toward
yellow-green. The tile on the left of the last panel is the $16\times16$
artwork the model drew; on its right it is repeated $3\times3$, with one copy
outlined.}
\label{fig:techniques}
\end{figure}

\textbf{Deliverable and judgment.} The drawing API offers geometric
primitives, polygons, B\'ezier paths, circles and rectangles, but the
deliverable differs from SVG generation. An SVG is a resolution-independent
description whose pixels a renderer decides, with anti-aliased edges and
freely blended colors; a PixelBench artwork is the native pixel matrix itself.
Every primitive lands on the native grid without anti-aliasing, only palette
colors exist, and the raters and the judge see the native pixels magnified. A
\texttt{circle} call therefore leaves a staircase of hard-edged pixels whose
cleanup is the model's decision, and the smaller the canvas, the less a
primitive helps: on a $16\times16$ canvas a circle of radius 3 is seven pixels
wide. The rubric asks that every pixel be placed deliberately at 32 pixels and
below (Appendix~\ref{app:prompts}).

\textbf{What the programs do.} We re-executed the submitted programs of all
1,200 artworks of Table~\ref{tab:main} with an instrumented engine that records
which call decided each pixel; 1,197 replays reproduce their artwork byte for
byte, and the other three are left out. Only 17\% of the programs call
\texttt{path}, and path-drawn pixels make up 3.2\% of the final images. On
canvases of 16 pixels and below no program uses a path, and 48\% of the
artworks are written out in the program as equal-length strings, one character
per pixel, which is how pixel art is drawn by hand. The submissions also show
techniques specific to the medium; Figure~\ref{fig:techniques} shows seven of
them, each from a generation task drawn from an empty canvas.

\textbf{Explicit pixels against primitives.} The direct-answer baseline of
Appendix~\ref{app:effort} forces an explicit decision for every pixel and
removes the primitives. On canvases of 32 pixels and below GPT-6 Astra then
matches its agent score ($80.8$ against $81.4$), and above 64 pixels it falls
behind ($69.1$ against $83.1$): on small canvases explicit pixel decisions
suffice, and on large ones primitives and programs help organize the
composition. The human rankings separate the two regimes as well. On the 28
generation tasks of 16 to 32 pixels, two random halves of the raters order the
twelve configurations alike (Spearman $\rho = 0.97$ on average), yet that order
differs from the order on the 42 tasks of 64 pixels and above ($\rho = 0.87$,
against $0.97$ and $0.99$ between rater halves within either band). The small
canvases measure pixel-level craft that the large ones do not.

\section{Creation behavior and failure modes}
\label{app:behavior}

\textbf{Rhythm of writing.} GPT-6 Astra and GPT-5.6 Sol write their
program about twice per task on average, at most 9 and 7 times, and usually
get the composition in one pass. Qwen3.8 Max, GLM-5.3 Flash, and Gemini 3.8
Flash write 11 to 13 times, revising as they go, and DeepSeek V4.1 Flash 15
times, using the full budget on 32 of its 100 artworks. More iteration does not
go with better artworks.

\textbf{Style of program.} GPT-6 Astra writes the shortest programs, a median
of 75 lines, and a third of them use paths; Gemini 3.8 Flash writes the
longest, a median of 442 lines, and almost never uses a path (1\%).

\textbf{Programming and drawing.} We compare the four configurations the raters
rank highest (GPT-6 Astra, Gemini 3.8 Flash, Claude Opus 5, Qwen3.8 Max) with
the four they rank lowest (Gemini 3.1 Pro Preview, DeepSeek V4 Pro, DeepSeek V4
Flash, Qwen3.5-397B-A17B). The lower four do not fail at programming: 5.4\% of
their program executions fail, against 4.9\%; 399 of their 400 artworks are
submitted formally, against 397; and among artworks whose programs never
failed, they trail by 4.3 places of mean human rank, the same gap as over all
artworks. They differ in drawing. On the generation tasks, rectangles, circles,
and ellipses produce 48\% of their drawn pixels, against 30\%, and 22\% of
their pixels lie inside flat single-color areas, against 12\%; within a task,
more flat area goes with a worse human rank (mean Spearman $\rho = -0.49$
$[-0.55, -0.42]$). In a review of 16 tasks drawn at random, the lower-ranked
artwork of each task's best and worst pair shows a misconstructed or unreadable
main shape in 10 cases, flat unshaded fills in 6, and weak composition in 4.
Dithering, outlines, and symmetry are equally common in both groups. Three of
the four lower configurations take no image input, yet GPT-6 Astra without
image observation keeps these properties ($15\%$ flat pixels, against $14\%$
with it).

\textbf{Common difficulties.} All configurations find characters, creatures,
and effects hardest (Section~\ref{sec:slices}), and editing depends on seeing
the original (Section~\ref{sec:variants}). Of the 1,200 artworks none is empty;
1,190 were submitted formally, and the other ten, all from runs that used the
full write budget, were recovered from the program's canvases
(Appendix~\ref{app:runs}).

\end{document}

%% file: math_commands.tex
\usepackage{amsmath,amsfonts,bm}

\def\eqref#1{equation~\ref{#1}}

\def\1{\bm{1}}

\DeclareMathAlphabet{\mathsfit}{\encodingdefault}{\sfdefault}{m}{sl}
\SetMathAlphabet{\mathsfit}{bold}{\encodingdefault}{\sfdefault}{bx}{n}

%% file: tables/table_main_b2.tex
\begin{tabular}{@{}lccccccccc@{}}
\toprule
& \multicolumn{3}{c}{All (100)} & \multicolumn{3}{c}{Generation (75)} & \multicolumn{3}{c}{Editing (25)} \\
\cmidrule(lr){2-4}\cmidrule(lr){5-7}\cmidrule(l){8-10}
Configuration & Score & \multicolumn{2}{c}{Rank $\downarrow$} & Score & \multicolumn{2}{c}{Rank $\downarrow$} & Score & \multicolumn{2}{c}{Rank $\downarrow$} \\
 & & Judge & Human & & Judge & Human & & Judge & Human \\
\midrule
GPT-6 Astra (high)              & \textbf{83.76} & \textbf{2.52} & \textbf{3.34} & \textbf{82.35} & \textbf{2.40} & \textbf{3.21} & \textbf{88.00} & \textbf{2.86} & \textbf{3.74} \\
Claude Opus 5 (low)             & 78.49 & 5.03 & 4.88 & 77.64 & 5.01 & 5.09 & 81.04 & 5.06 & 4.25 \\
Qwen3.8 Max                     & 78.04 & 4.72 & 5.13 & 77.35 & 4.54 & 5.00 & 80.12 & 5.26 & 5.49 \\
Gemini 3.8 Flash (high)         & 77.35 & 4.69 & 4.27 & 77.87 & 4.45 & 3.82 & 75.80 & 5.40 & 5.62 \\
GPT-5.6 Sol (medium)            & 76.92 & 5.51 & 5.91 & 76.29 & 5.47 & 5.82 & 78.80 & 5.62 & 6.21 \\
DeepSeek V4.1 Flash (high)      & 75.70 & 5.41 & 6.45 & 76.27 & 5.21 & 6.40 & 74.00 & 6.00 & 6.60 \\
GLM-5.3 Flash (max)             & 74.98 & 6.11 & 6.53 & 75.44 & 6.04 & 6.48 & 73.60 & 6.32 & 6.70 \\
Kimi K3 (max)                   & 74.45 & 6.61 & 6.42 & 74.23 & 6.79 & 6.40 & 75.12 & 6.08 & 6.45 \\
Gemini 3.1 Pro Preview (high)   & 69.51 & 8.12 & 6.78 & 68.13 & 8.47 & 7.11 & 73.64 & 7.08 & 5.76 \\
DeepSeek V4 Pro (high)          & 67.41 & 8.88 & 8.74 & 67.73 & 9.09 & 8.90 & 66.44 & 8.24 & 8.26 \\
DeepSeek V4 Flash (high)        & 65.52 & 9.29 & 9.08 & 66.88 & 9.36 & 9.23 & 61.44 & 9.08 & 8.63 \\
Qwen3.5-397B-A17B               & 55.60 & 11.12 & 10.47 & 57.89 & 11.16 & 10.53 & 48.72 & 11.00 & 10.29 \\
\bottomrule
\end{tabular}

%% file: tables/table_variants_b2.tex
\begin{tabular}{@{}llccc@{}}
\toprule
Model & Setting & All (100) & Generation (75) & Editing (25) \\
\midrule
GPT-6 Astra        & high (Table~\ref{tab:main})      & 83.8 & 82.3 & 88.0 \\
                   & low                              & 81.8 & 80.5 & 85.8 \\
                   & medium                           & 82.4 & 80.8 & 87.4 \\
                   & xhigh                            & 84.7 & 83.6 & 87.9 \\
                   & no image observation             & 79.9 & 80.5 & 77.8 \\
                   & direct answer                    & 76.9 & 76.5 & 78.1 \\
                   & creation runs 2 / 3              & 83.0 / 81.7 & 81.8 / 80.3 & 86.5 / 85.8 \\
\midrule
GPT-5.6 Sol        & medium (Table~\ref{tab:main})    & 76.9 & 76.3 & 78.8 \\
                   & creation runs 2 / 3              & 77.4 / 76.7 & 76.7 / 76.1 & 79.3 / 78.5 \\
\midrule
DeepSeek V4 Flash  & high (Table~\ref{tab:main})      & 65.5 & 66.9 & 61.4 \\
                   & no reasoning                     & 62.4 & 64.9 & 54.8 \\
                   & low                              & 61.0 & 62.9 & 55.2 \\
                   & max                              & 66.1 & 68.0 & 60.5 \\
                   & direct answer                    & 18.6 & 17.1 & 23.0 \\
                   & creation runs 2 / 3              & 64.0 / 65.4 & 66.9 / 66.9 & 55.2 / 60.6 \\
\midrule
GLM-5.3 Flash      & max (Table~\ref{tab:main})       & 75.0 & 75.4 & 73.6 \\
                   & low                              & 62.6 & 65.1 & 55.1 \\
                   & high                             & 69.8 & 72.2 & 62.8 \\
\bottomrule
\end{tabular}

%% file: figures/gallery_pages_b2.tex
\begin{figure}[p]
\centering
\includegraphics[page=1,width=0.99\textwidth]{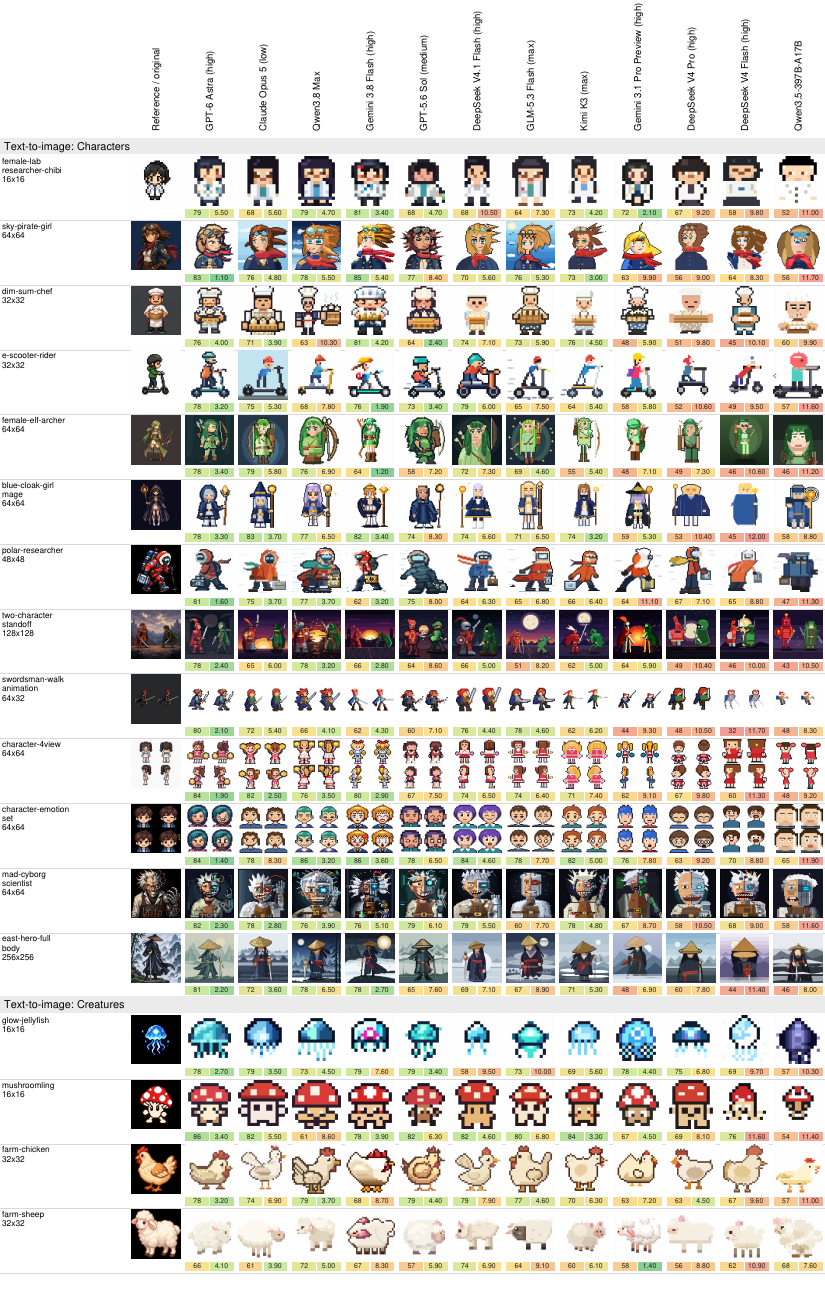}
\caption{Full gallery, part 1 of 6.}
\label{fig:gallery1}
\end{figure}

\begin{figure}[p]
\centering
\includegraphics[page=2,width=0.99\textwidth]{figures/gallery_b2}
\caption{Full gallery, part 2 of 6.}
\label{fig:gallery2}
\end{figure}

\begin{figure}[p]
\centering
\includegraphics[page=3,width=0.99\textwidth]{figures/gallery_b2}
\caption{Full gallery, part 3 of 6.}
\label{fig:gallery3}
\end{figure}

\begin{figure}[p]
\centering
\includegraphics[page=4,width=0.99\textwidth]{figures/gallery_b2}
\caption{Full gallery, part 4 of 6.}
\label{fig:gallery4}
\end{figure}

\begin{figure}[p]
\centering
\includegraphics[page=5,width=0.99\textwidth]{figures/gallery_b2}
\caption{Full gallery, part 5 of 6.}
\label{fig:gallery5}
\end{figure}

\begin{figure}[p]
\centering
\includegraphics[page=6,width=0.99\textwidth]{figures/gallery_b2}
\caption{Full gallery, part 6 of 6.}
\label{fig:gallery6}
\end{figure}

%% file: figures/tasks_appendix_b2.tex
\subsection{Text-to-image tasks}
\label{app:tasks-gen}

\begin{multicols}{3}\raggedright
\taskgroup{Characters (13)}
\taskentry{Female Lab Researcher Chibi}{female-lab-researcher-chibi}{$16\times16$, colors $\leq 86$}{pixel art, 16x16 resolution, RPG game, two-heads-tall chibi, female, lab researcher, white lab coat, black hair, brown eyes}
\taskentry{Sky Pirate Girl}{sky-pirate-girl}{$64\times64$, colors $\leq 86$}{pixel art, 64x64 resolution, anime style, character portrait, young sky pirate girl, goggles, short navy coat, red scarf, three-quarter view, confident expression, wind-swept hair, clear silhouette}
\taskentry{Dim Sum Chef}{dim-sum-chef}{$32\times32$, colors $\leq 86$}{pixel art, 32x32 resolution, dim sum chef, front-facing character sprite, flour-dusted apron, holding bamboo steamer with buns, hands steamer food clothing clearly separated}
\taskentry{Electric Scooter Rider}{e-scooter-rider}{$32\times32$, colors $\leq 86$}{pixel art, 32x32 resolution, little person riding an electric scooter, side view riding right, helmet, e-scooter with standing deck and handlebars, readable riding pose}
\taskentry{Female Elf Archer}{female-elf-archer}{$64\times64$, colors $\leq 86$}{pixel art, 64x64 resolution, RPG character portrait, female elf archer, long green hair, pointed ears, light green outfit, bow and arrows on back, standing pose, three-quarter view}
\taskentry{Blue-Cloak Girl Mage}{blue-cloak-girl-mage}{$64\times64$, colors $\leq 86$}{pixel art, 64x64 resolution, anime style, full-body character, girl mage, relaxed pose, dark blue cloak, white boots, glowing amber staff, strong silhouette, simple readable costume, front-facing, RPG character sprite}
\taskentry{Polar Researcher}{polar-researcher}{$48\times48$, colors $\leq 86$}{pixel art, 48x48 resolution, polar researcher character sprite, transparent background, heavy cold-weather clothing, carrying sample case, pale blue goggle reflections, forward-leaning pose suggesting strong wind}
\taskentry{Two-Character Standoff}{two-character-standoff}{$128\times128$, colors $\leq 86$}{pixel art, 128x128 resolution, anime style, two rival characters, dramatic standoff, red-armored knight left, green-cloaked rogue right, clear gap between them, readable silhouettes}
\taskentry{Swordsman Walk Animation}{swordsman-walk-animation}{$64\times32$ (two $32\times32$ frames), colors $\leq 86$}{pixel art, 64x32 resolution, 16-bit RPG style, red-haired swordsman, walk animation, 2-frame sequence, side view facing right, long sword, dark blue cape, smooth natural motion, clear frame transitions, single frame 32x32}
\taskentry{Character Four Views}{character-4view}{$64\times64$, colors $\leq 86$}{pixel art, 64x64 resolution, same character in four views, 2x2 grid, top row front and back, bottom row left and right side, same character and outfit in all four views, girl, cheerleader, dark outline, transparent background}
\taskentry{Character Emotion Set}{character-emotion-set}{$64\times64$, colors $\leq 86$}{pixel art, 64x64 resolution, big head portraits of the same character in four emotions, 2x2 grid, calm, smiling, angry, surprised, same character with same hair and outfit in all four, expressive and clearly readable, dark outline, transparent background}
\taskentry{Mad Cyborg Scientist}{mad-cyborg-scientist}{$64\times64$, colors $\leq 86$}{pixel art, 64x64 resolution, RPG character portrait, mad cyborg scientist, left side human face, right side metal robotic head, one glowing cybernetic eye, exposed metal robotic arm, prosthetic hand with cables, messy white hair, brain electrode plugs, dirty lab coat, leather apron, tool belt, maniacal grin, strong light contrast}
\taskentry{Eastern Hero Full Body}{east-hero-full-body}{$256\times256$, colors $\leq 120$}{pixel art, 256x256 resolution, full body character portrait, oriental wuxia swordsman, wide bamboo conical hat (douli) tilted down covering the upper half of the face, long flowing dark robe with sash, loose hair, one hand resting on a straight jian sword, misty mountain backdrop, three-quarter view, heroic stance, flowing fabric details, dark outline}
\taskgroup{Creatures (10)}
\taskentry{Glow Jellyfish}{glow-jellyfish}{$16\times16$, colors $\leq 86$}{pixel art, 16x16 resolution, RPG enemy, small glowing jellyfish in deep water, translucent glowing bell, short tentacles, dark outline, transparent background}
\taskentry{Mushroomling}{mushroomling}{$16\times16$, colors $\leq 4$}{pixel art, 16x16 resolution, at most 4 colors, RPG enemy, small mushroom monster, red cap with white spots, round eyes, stubby arms and legs standing, dark outline, transparent background}
\taskentry{Chicken (Farm Style)}{farm-chicken}{$32\times32$, colors $\leq 86$}{pixel art, 32x32 resolution, Stardew Valley style, cozy farm animal, a plump chicken, gentle side view, soft rounded shapes, warm muted palette, transparent background}
\taskentry{Sheep (Farm Style)}{farm-sheep}{$32\times32$, colors $\leq 86$}{pixel art, 32x32 resolution, Stardew Valley style, cozy farm animal, fluffy white sheep with pale pink ears, side view, soft rounded shapes, gentle flat shading, no harsh outline, warm muted palette, transparent background}
\taskentry{Forest Slime Monster}{forest-slime-monster}{$32\times32$, colors $\leq 86$}{pixel art, 32x32 resolution, cute forest slime monster, moss, tiny antlers, glowing eyes, small red mushroom on back, front-facing, RPG enemy sprite, compact silhouette, readable expression}
\taskentry{Monarch Butterfly}{monarch-butterfly}{$32\times32$, colors $\leq 86$}{pixel art, 32x32 resolution, monarch butterfly, transparent background}
\taskentry{Skeleton Grunt}{skeleton-grunt}{$32\times32$, colors $\leq 86$}{pixel art, 32x32 resolution, RPG enemy, skeleton grunt, short sword and small wooden shield, slightly hunched pose, dark joint shading, transparent background}
\taskentry{Dragon Hatchling}{dragon-hatchling}{$64\times64$, colors $\leq 86$}{pixel art, 64x64 resolution, RPG enemy, young western dragon hatchling, half-spread wings, scaled body with pale belly, small horns, spiked tail, low growl with open mouth, strong silhouette, transparent background}
\taskentry{Void Eye}{void-eye}{$64\times64$, colors $\leq 86$}{pixel art, 64x64 resolution, RPG boss, giant floating single-eye monster, deep purple iris, faintly glowing pupil, short tentacles, hovering with soft shadow, ominous mood, strong contrast, readable silhouette, transparent background}
\taskentry{Ancient Treant Boss}{ancient-treant-boss}{$256\times256$, colors $\leq 120$}{pixel art, 256x256 resolution, boss creature, massive ancient treant, towering tree-like humanoid made of gnarled bark, glowing green eyes in the trunk, moss and hanging vines, branch arms with claw-like twigs, roots rising from the ground, forest clearing with god rays, menacing towering pose, highly detailed, dark outline}
\taskgroup{Scenes and architecture (16)}
\taskentry{Summer Seaside}{summer-seaside}{$128\times128$, colors $\leq 86$}{pixel art, 128x128 resolution, summer seaside beach scene, clear blue sky with soft white clouds, bright sun, sea layered from teal near shore to deep blue at the horizon, white foam waves rolling onto golden sand, red-and-white striped beach umbrella, striped towel and flip-flops, small white sailboat far out, two gulls flying, bright fresh summer mood, strong daylight}
\taskentry{Neighborhood Laundromat}{neighborhood-laundromat}{$64\times64$, colors $\leq 86$}{pixel art, 64x64 resolution, small self-service laundromat interior, two rows of front-loading washing machines, laundry basket, folding table, wall clock, cool white lighting, individual machines visually separated}
\taskentry{Mountain Village at Sunrise}{mountain-village-sunrise}{$128\times128$, colors $\leq 86$}{pixel art, 128x128 resolution, mountain village at sunrise, three wooden houses, chimney smoke, stone path, pine trees, distant mountains, layered depth, clear focal point center}
\taskentry{Red-Roofed Forge House}{red-roofed-forge-house}{$128\times128$, colors $\leq 86$}{pixel art, 128x128 resolution, isometric building, red tiled roof, furnace glow, wooden sign, stacked barrels, small chimney, clean 45-degree geometry, no perspective distortion, strategy game tileset, RTS game style}
\taskentry{Rice Terraces During Harvest}{rice-terraces-during-harvest}{$64\times64$, colors $\leq 86$}{pixel art, 64x64 resolution, top-down golden rice terraces following mountain contours, harvest season, two farmers on narrow ridges, simple village houses in distance, layered values for depth}
\taskentry{Sunken Shipwreck}{sunken-shipwreck}{$64\times64$, colors $\leq 86$}{pixel art, 64x64 resolution, underwater scene, sunken shipwreck, half-buried wooden hull in sand, corroded ship body, school of fish swimming past, coral and seaweed accents}
\taskentry{Tavern Interior}{tavern-interior}{$64\times64$, colors $\leq 86$}{pixel art, 64x64 resolution, cozy indoor tavern, wooden bar counter, shelf of bottles behind the bar, bar stools, warm yellow hanging lamps, glasses and wooden barrels, smoky atmosphere, warm nighttime lighting, dark outline}
\taskentry{University Laboratory Workbench}{university-laboratory-workbench}{$64\times64$, colors $\leq 86$}{pixel art, 64x64 resolution, laboratory workbench, microscope, test-tube rack, dropper, labeled bottles, open laboratory notebook, believable relative sizes, distinguish objects by shape rather than unreadable tiny text}
\taskentry{Castle Title Screen}{castle-title-screen}{$128\times128$, colors $\leq 86$}{pixel art, 128x128 resolution, fantasy RPG title screen, castle on seaside cliff, sunset, two towers with flags, large orange sun, layered clouds, dark rocks foreground, wide cinematic composition}
\taskentry{Cyberpunk Alley}{cyberpunk-alley}{$128\times128$, colors $\leq 86$}{pixel art, 128x128 resolution, cyberpunk alley, rainy night, neon shop signs, reflective puddles, power cables, vending machines, one person with transparent umbrella, strong light contrast, readable silhouettes}
\taskentry{Poolcore Water Slide}{poolcore-water-slide}{$128\times128$, colors $\leq 86$}{pixel art, 128x128 resolution, poolcore liminal space, backrooms-style indoor pool hall, vast empty indoor pool room, crystal clear pale-blue water with gentle ripples and reflections, a curving white water slide descending into the water, floor and walls covered in pale-blue-and-white grid mosaic tiles, tiled columns, bright fluorescent ceiling lights casting even cool light, a girl seen from behind floating in a yellow swim ring in the middle of the pool, quiet dreamy uncanny atmosphere}
\taskentry{Convenience Store on a Rainy Night}{convenience-store-on-a-rainy-night}{$64\times64$, colors $\leq 86$}{pixel art, 64x64 resolution, small convenience store on rainy night, bright interior light through glass windows, umbrella rack, parked bicycle, reflective puddles, person in raincoat entering, no real brand names or logos}
\taskentry{Subway Platform}{subway-platform}{$64\times64$, colors $\leq 86$}{pixel art, 64x64 resolution, subway platform side view, train arriving, platform screen doors, bench, station signs without brand names, three passengers with different poses, horizontal composition conveying speed and depth, readable silhouettes}
\taskentry{Magic Forest Cottage}{magic-forest-cottage}{$128\times128$, colors $\leq 86$}{pixel art, 128x128 resolution, magic forest cottage, deep purple night sky, full moon, mist, old wood-and-stone cottage, steep mossy roof, warm orange window light, stone well, wooden winch, stone-slab path, twisted ancient trees, dark red brown purple leaves, fairy-tale mood, warm-cold contrast}
\taskentry{Offshore Wind Turbine Maintenance}{offshore-wind-turbine-maintenance}{$64\times64$, colors $\leq 86$}{pixel art, 64x64 resolution, offshore wind turbine maintenance platform, technician in safety equipment climbing ladder, two additional turbines in distance, believable human scale, clearly separate steel structure ocean waves sky}
\taskentry{Ancient Dragon Battle}{ancient-dragon-battle}{$256\times256$, colors $\leq 140$}{pixel art, 256x256 resolution, epic fantasy battlefield scene, ancient dragon flying above a burning castle, knights with spears and shields on ramparts, arrows flying, smoke and embers, dramatic sunset sky, siege towers, soldiers in formation, distant mountains, rich detail, cinematic composition, dark outline}
\taskgroup{Objects (21)}
\taskentry{Magnifying Glass}{magnifying-glass}{$16\times16$, colors $\leq 86$}{pixel art, 16x16 resolution, detective game, magnifying glass icon}
\taskentry{Gold Star Coin}{gold-star-coin}{$16\times16$, colors $\leq 6$}{pixel art, 16x16 resolution, at most 6 colors, gold coin, star symbol in the middle}
\taskentry{Wooden Key}{wooden-key}{$16\times16$, colors $\leq 6$}{pixel art, 16x16 resolution, at most 6 colors, detective game, wooden key}
\taskentry{Leather Spellbook}{leather-spellbook}{$32\times32$, colors $\leq 8$}{pixel art, 32x32 resolution, at most 8 colors, RPG item portrait, book, closed spellbook, leather cover, metal corners, spine, dark outline, transparent background}
\taskentry{USB Flash Drive}{usb-flash-drive}{$32\times32$, colors $\leq 86$}{pixel art, 32x32 resolution, RPG item portrait, USB flash drive, metal casing, USB connector, keychain strap, dark outline, transparent background}
\taskentry{Retro Film Camera}{retro-film-camera}{$32\times32$, colors $\leq 86$}{pixel art, 32x32 resolution, RPG item portrait, camera, retro film camera, lens, shutter button, leather texture, dark outline, transparent background}
\taskentry{Green Pear}{green-pear}{$32\times32$, colors $\leq 7$}{pixel art, 32x32 resolution, at most 7 colors, green pear}
\taskentry{Bamboo Steamer}{bamboo-steamer}{$32\times32$, colors $\leq 86$}{pixel art, 32x32 resolution, bamboo steamer, transparent background, lid partially open, three clearly shaped dumplings inside, small rising steam}
\taskentry{Laboratory Microscope Icon}{laboratory-microscope-icon}{$32\times32$, colors $\leq 86$}{pixel art, 32x32 resolution, laboratory microscope icon, transparent background, three-quarter view, recognizable eyepiece body tube stage objective base}
\taskentry{Portable Cassette Player}{portable-cassette-player}{$32\times32$, colors $\leq 86$}{pixel art, 32x32 resolution, portable cassette player, front-facing icon, transparent background, transparent cassette window two tape reels, playback controls volume control headphone cable}
\taskentry{Chainmail Armor}{chainmail-armor}{$32\times32$, colors $\leq 86$}{pixel art, 32x32 resolution, RPG armor, chainmail, interlocking metal rings, dark outline, transparent background}
\taskentry{Magic Staff}{magic-staff}{$48\times48$, colors $\leq 86$}{pixel art, 48x48 resolution, RPG wooden magic staff, curved head, purple particle effects, transparent background}
\taskentry{City Bus}{city-bus}{$64\times32$, colors $\leq 86$}{pixel art, 64x32 resolution, modern city bus, side-view sprite, transparent background, front rear doors, row of windows, two wheels, headlights, simplified electronic route display, no real company branding}
\taskentry{Vintage Sewing Machine}{vintage-sewing-machine}{$48\times48$, colors $\leq 86$}{pixel art, 48x48 resolution, mechanical sewing machine, transparent background, slightly angled side view, handwheel needle bar presser foot base red fabric}
\taskentry{Family Dining Table}{dining-table}{$64\times64$, colors $\leq 86$}{pixel art, 64x64 resolution, furniture item sprite, family dining table, wooden table, four chairs around it, a bowl of fruit on the table, cozy family atmosphere, top-down or front view, dark outline}
\taskentry{Fast Food Combo}{fast-food-combo}{$64\times64$, colors $\leq 86$}{pixel art, 64x64 resolution, 45 degree top-down view of a fast food combo meal on a tray, cheeseburger with sesame bun lettuce and melted cheese, golden french fries in a red carton, crispy chicken nuggets, cup of cola with ice and straw, dipping sauce cups, appetizing food styling, clean details, dark outline}
\taskentry{Weapon Set Icons}{weapon-set-icons}{$64\times64$, colors $\leq 86$}{pixel art, 64x64 resolution, weapon icon set, 2x2 grid, sword, bow, shield, battle axe, four weapons in one consistent style, dark outline, transparent background}
\taskentry{Realistic Heart}{realistic-heart}{$64\times64$, colors $\leq 86$}{pixel art, 64x64 resolution, realistic style, heart}
\taskentry{Motorcycle}{motorcycle}{$128\times128$, colors $\leq 86$}{pixel art, 128x128 resolution, sports motorcycle, side profile, facing right, streamlined, black and dark gray body, silver gray metal parts, red accents, tilted small black windscreen, wide dark gray fairing, silver front fork, thick black tires, silver tubular frame, round silver gray engine cover, upswept tail, red tail light, gray hub white spokes, transparent background}
\taskentry{Open Pocket Watch}{open-pocket-watch}{$64\times64$, colors $\leq 86$}{pixel art, 64x64 resolution, realistic style, open mechanical pocket watch, lid open, intricate gear train clearly visible, golden metal movement, meshing gears, dial with tick marks, hands, watch chain, macro view, metallic texture and reflections, strong light contrast}
\taskentry{Giant Arcane Orrery}{arcane-instrument-giant}{$256\times256$, colors $\leq 140$}{pixel art, 256x256 resolution, huge complex arcane instrument, massive brass astrolabe-orrery with concentric engraved rings, orbiting crystal spheres, glowing runes along the rings, central radiant core, intricate gears and alchemical dials, floating over a stone pedestal in a dark hall, extremely detailed machinery, dramatic lighting, dark outline}
\taskgroup{Icons and interface (7)}
\taskentry{RTS Factory Icon}{rts-factory-icon}{$32\times32$, colors $\leq 86$}{pixel art, 32x32 resolution, RTS game, factory building icon, chimney, flat roof, brick walls, vents, clean silhouette, centered composition, strategy game tileset}
\taskentry{Battery Status Set}{battery-status-set}{$64\times64$, colors $\leq 86$}{pixel art, 64x64 resolution, four battery icons in one horizontal row, 100\% charge 60\% 20\% active charging, identical battery outlines, clearly distinguish four fill states}
\taskentry{Compass}{item-icon}{$32\times32$, colors $\leq 86$}{pixel art, 32x32 resolution, inventory icon, ancient golden compass, blue crystal center, top-down view, thick dark outline, centered composition, game UI icon}
\taskentry{Typography}{typography}{$64\times64$, colors $\leq 86$}{Create a pixel-art game title screen featuring the words ``PIXEL QUEST''. Design clear, readable pixel lettering, using letter size, spacing, and arrangement to establish a strong visual hierarchy. You may add outlines, shadows, or small decorative elements, while keeping the lettering as the main focus.}
\taskentry{Chart Layout}{chart-layout}{$128\times64$, colors $\leq 86$}{pixel art, 128x64 resolution, bar chart, fruit data, apple 10, banana 5, orange 7, lemon 14, grape 2, color bars by fruit, mini pixel art icons beside chart instead of text}
\taskentry{Skill Icon Set}{skill-icon-set}{$64\times64$, colors $\leq 86$}{pixel art, 64x64 resolution, skill icon set, 2x2 grid, fireball, ice shield, healing, lightning, each icon 32x32, four skills in one consistent style, dark outline, transparent background}
\taskentry{Public Transportation Icon Set}{public-transportation-icon-set}{$64\times64$, colors $\leq 86$}{pixel art, 64x64 resolution, consistent 2x2 grid of public transportation icons, bus, subway, bicycle, pedestrian, consistent framing icon scale line weight palette background}
\taskgroup{Materials (5)}
\taskentry{Dirt Material (Minecraft Style)}{minecraft-dirt-material}{$16\times16$, colors $\leq 10$}{pixel art, 16x16 resolution, at most 10 colors, game material texture, Minecraft style, dirt block, seamless tile, granular dirt texture}
\taskentry{Stone Material (Minecraft Style)}{minecraft-stone-material}{$16\times16$, colors $\leq 10$}{pixel art, 16x16 resolution, at most 10 colors, game material texture, Minecraft style, stone block, seamless tile, cracked stone texture}
\taskentry{Grass Material (RPG Style)}{rpg-grass-material}{$16\times16$, colors $\leq 86$}{pixel art, 16x16 resolution, RPG game style, grass material tile, seamless tiling, interleaved dark and light green grass speckles, a few tiny yellow flowers, small darker grass tufts, sparse bare-dirt patches}
\taskentry{Wooden Planks Material (RPG Style)}{rpg-wooden-planks-material}{$16\times16$, colors $\leq 86$}{pixel art, 16x16 resolution, RPG game style, wooden planks material tile, seamless tiling, several vertical boards side by side, natural wood-grain lines, dark seam grooves between planks, alternating light and dark board ends, a few wood knots}
\taskentry{Seamless Carpet Pattern}{seamless-carpet-pattern}{$32\times32$, colors $\leq 86$}{pixel art, 32x32 resolution, seamless carpet pattern, warm geometric motifs, diamonds with floral border, woven texture, seamless tiling in all directions, left right edges connect exactly}
\taskgroup{Effects (3)}
\taskentry{Slash Effect}{slash-effect}{$48\times48$, colors $\leq 86$}{pixel art, 48x48 resolution, slash effect, 2 frames, frame 1 a small white arc slash, frame 2 a larger arc slash with a trailing streak, transparent background}
\taskentry{Teleport Portal}{teleport-portal}{$48\times48$, colors $\leq 86$}{pixel art, 48x48 resolution, teleport portal, 2 frames, frame 1 a purple energy vortex, frame 2 the vortex rotated with a glowing halo, transparent background}
\taskentry{Explosion Effect}{explosion-effect}{$64\times64$, colors $\leq 86$}{pixel art, 64x64 resolution, explosion effect, 2 frames, frame 1 a bright yellow fireball at center, frame 2 an expanding orange-red blast with sparks, transparent background}
\end{multicols}

\subsection{Editing tasks}
\label{app:tasks-edit}

\noindent Instruction editing (13), animation (10), multiview (2), in this order.\par\smallskip
\begin{tcbraster}[raster columns=3, raster equal height=rows, raster column skip=4pt, raster row skip=4pt]
\begin{editcardv}{Mage and Tower}{mage-and-tower}{figures/tasks/big-144-256}
Instruction editing, single frame \quad Input $144\times256$\par
\taskprompt{Let a layer of vines cover part of the tower's walls. Add a cloaked mage holding a glowing staff at the archway entrance of the tower; he is walking into the tower (facing away from the viewer, heading through the archway).}\par
\tasksource{Pixiv, artwork 123659418.}{Not stated}
\end{editcardv}
\begin{editcardv}{Vending Machine in a Thunderstorm}{thunderstorm-vending-machine}{figures/tasks/big-170-170}
Instruction editing, single frame \quad Input $170\times170$\par
\taskprompt{Remove the character from the scene (and fill in the parts it was covering, such as the vending machine). Change the weather to a thunderstorm: make the ground wet with clinging water droplets, and render the rain with semi-transparent droplets. Add lightning in the background.}\par
\tasksource{Pixiv, artwork 123439931.}{Not stated}
\end{editcardv}
\begin{editcardv}{Mountains at Noon}{mountains-at-noon}{figures/tasks/big-250-147}
Instruction editing, single frame \quad Input $250\times147$\par
\taskprompt{Set the time to noon on a clear day. Note: because it is noon, the sun should not appear in the picture.}\par
\tasksource{OpenGameArt, ``Mountain at Dusk Background'' by ansimuz.}{CC0}
\end{editcardv}
\begin{editcardv}{Demon King's Castle}{demon-king-castle}{figures/tasks/big-256-128}
Instruction editing, single frame \quad Input $256\times128$\par
\taskprompt{Edit the pixel art based on the original image: Change the theme of the whole picture: turn the ``City of Hope'' into the Demon King's castle beneath a hanging blood moon.}\par
\tasksource{Pixiv, artwork 123985822.}{Not stated}
\end{editcardv}
\begin{editcardv}{Potion Bottle}{bottle-potion}{figures/tasks/bottle-potion}
Instruction editing, single frame \quad Input $64\times64$\par
\taskprompt{Turn the water in the bottle into a bubbling purple magic potion.}\par
\taskorigin{Drawn by the authors.}
\end{editcardv}
\begin{editcardv}{Durian}{durian-segment}{figures/tasks/durian-segment}
Instruction editing, single frame \quad Input $64\times64$\par
\taskprompt{Remove the bottom segment of durian flesh so the rind underneath is exposed; keep the other two pieces unchanged.}\par
\tasksource{Pixel Beads Art (\texttt{pixel-beads.com}), durian bead pattern.}{Not stated}
\end{editcardv}
\begin{editcardv}{Fountain Water Flow}{fountain-flow}{figures/tasks/fountain-flow}
Instruction editing, single frame \quad Input $64\times76$\par
\taskprompt{The fountain is currently in a waterless state. Change it into a fountain that has water while it is running.}\par
\tasksource{OpenGameArt.}{Not stated}
\end{editcardv}
\begin{editcardv}{Knife}{knife}{figures/tasks/knife}
Instruction editing, single frame \quad Input $93\times128$\par
\taskprompt{Pose this knife thrusting to the right with its cutting edge facing downward, leaving semi-transparent afterimages and a white slash effect.}\par
\tasksource{OpenGameArt, ``Knife'' by kotnaszynce; frame 1 of \texttt{nurz.gif}.}{CC0}
\end{editcardv}
\begin{editcardv}{Melting Ice Cream}{melting-icecream}{figures/tasks/melting-icecream}
Instruction editing, single frame \quad Input $64\times64$\par
\taskprompt{Make the ice cream's chocolate melt and drip down the cone.}\par
\taskorigin{Drawn by the authors.}
\end{editcardv}
\begin{editcardv}{Motorcycle on a Country Road}{motorcycle-road}{figures/tasks/motorcycle-road}
Instruction editing, single frame \quad Input $64\times64$\par
\taskprompt{Put the motorcycle onto a country road.}\par
\tasksource{Pixel Beads Art (\texttt{pixel-beads.com}), motorcycle bead pattern.}{Not stated}
\end{editcardv}
\begin{editcardv}{Snow on the Pine}{snowy-pine}{figures/tasks/snowy-pine}
Instruction editing, single frame \quad Input $82\times135$\par
\taskprompt{Cover the pine tree with a layer of snow.}\par
\tasksource{OpenGameArt, ``Pine Tree Pack'' by KnoblePersona.}{CC BY 3.0}
\end{editcardv}
\begin{editcardv}{Sword into Staff}{sword-to-staff}{figures/tasks/sword-to-staff}
Instruction editing, single frame \quad Input $64\times64$\par
\taskprompt{Change the sword in the adventurer's hand into a staff.}\par
\taskorigin{Drawn by the authors.}
\end{editcardv}
\begin{editcardv}{Yellow Bird on the Right}{yellow-bird-right}{figures/tasks/yellow-bird-right}
Instruction editing, single frame \quad Input $64\times64$\par
\taskprompt{Change the right-hand one of the two parrots into a yellow parrot.}\par
\tasksource{Pixel Beads Art (\texttt{pixel-beads.com}), parrot bead pattern by cglow.}{Not stated}
\end{editcardv}
\begin{editcardv}{Bat Flapping Wings}{bat-flapping-wings}{figures/tasks/bat-flapping-wings}
Animation, 2 frames \quad Input $27\times22$\par
\taskprompt{2 frames. Frame 1 keeps the original picture content (the canvas may be changed); frame 2 shows the bat flapping its wings: the drooping wings open up into a spread, raised position.}\par
\tasksource{OpenGameArt.}{Not stated}
\end{editcardv}
\begin{editcardv}{Bunny Hopping}{bunny-hop}{figures/tasks/bunny-hop}
Animation, 2 frames \quad Input $26\times28$\par
\taskprompt{2 frames. Frame 1 keeps the original picture content (the canvas may be changed); frame 2 shows the rabbit hopping forward, caught in a mid-leap pose.}\par
\tasksource{OpenGameArt.}{Not stated}
\end{editcardv}
\begin{editcardv}{Demon Cyclops Hit}{demon-cyclop-hit}{figures/tasks/demon-cyclop-hit}
Animation, 3 frames \quad Input $37\times33$\par
\taskprompt{3 frames. Frame 1 keeps the original picture content (the canvas may be changed); frame 2 shows the monster being hit: the foot on the left of the image (the monster's right foot) lifts off the ground, the whole body leans back, its single eye rolls upward and the mouth opens - the recoil pose; frame 3 shows the blow landing, with a red filter added over the whole monster to show the impact.}\par
\tasksource{Ninja Adventure Asset Pack by Pixel-boy and AAA (itch.io), Actor/Boss/DemonCyclop2; frames 1, 4, 2 of \texttt{HitPreview.gif}.}{CC0 1.0}
\end{editcardv}
\begin{editcardv}{Draw Sword Slash}{draw-sword-slash}{figures/tasks/draw-sword-slash}
Animation, 2 frames \quad Input $62\times72$\par
\taskprompt{2 frames. Frame 1 keeps the original picture content (the canvas may be changed); frame 2 shows the completed drawn-sword slash. This character is left-handed: the sword is held in the LEFT hand (the hand gripping the blade in the image) and the right hand does not hold it. The slash sweeps from the lower left up to the upper right - a left-to-right motion - leaving a black slash trail. The motion ends on the completed slash.}\par
\tasksource{CraftPix free asset.}{CraftPix free-asset license}
\end{editcardv}
\begin{editcardv}{Elf Archer Attack}{elf-archer-attack}{figures/tasks/elf-archer-attack}
Animation, 2 frames \quad Input $35\times47$\par
\taskprompt{2 frames. Frame 1 keeps the original picture content (the canvas may be changed); frame 2 shows the elf archer drawing the bow to its maximum - the bow fully open with the whole pose shifted into a proper drawn-bow stance. No arrow.}\par
\tasksource{Battle for Wesnoth core unit sprites, \texttt{sharpshooter-bow} attack animation, frames 1 and 4.}{GPL v2+ or CC BY-SA 4.0, as for the corresponding Wesnoth asset}
\end{editcardv}
\begin{editcardv}{Fist Swing}{fist-swing}{figures/tasks/fist-swing}
Animation, 2 frames \quad Input $29\times44$\par
\taskprompt{2 frames. Frame 1 keeps the original picture content (the canvas may be changed); frame 2 shows the character throwing a light punch with the left fist (the arm that extends horizontally to the right of the image).}\par
\tasksource{OpenGameArt.}{Not stated}
\end{editcardv}
\begin{editcardv}{Flag Waving}{flag-wave}{figures/tasks/flag-wave}
Animation, 3 frames \quad Input $48\times56$\par
\taskprompt{3 frames. Frame 1 keeps the original picture content (the canvas may be changed); frames 2 and 3 continue the looping flag-waving animation (the fly end swings from its highest to its lowest point).}\par
\tasksource{Ninja Adventure Asset Pack by Pixel-boy and AAA (itch.io), Backgrounds/Animated/Flag; frames 1--3 of \texttt{Preview.gif}.}{CC0 1.0}
\end{editcardv}
\begin{editcardv}{Paladin Attack}{paladin-attack}{figures/tasks/paladin-attack}
Animation, 2 frames \quad Input $43\times61$\par
\taskprompt{2 frames. Frame 1 keeps the original picture content (the canvas may be changed); frame 2 shows the paladin swinging the weapon in hand down in a heavy attack. A white motion trail streams out behind the weapon. The footing changes as well: as the weapon comes down, the left foot also steps forward to put weight behind the blow.}\par
\tasksource{Battle for Wesnoth core unit sprites, \texttt{royal-warrior} attack animation, frames 1 and 2.}{GPL v2+ or CC BY-SA 4.0, as for the corresponding Wesnoth asset}
\end{editcardv}
\begin{editcardv}{Release Flames}{release-flames}{figures/tasks/release-flames}
Animation, 2 frames \quad Input $47\times63$\par
\taskprompt{2 frames. Frame 1 keeps the original picture content (the canvas may be changed); frame 2 shows the character thrusting a fire spell forward: the flame flies forward with a fiery trail. The pose changes too - the left foot steps forward and the body leans in, pushing the flame out.}\par
\tasksource{CraftPix free asset.}{CraftPix free-asset license}
\end{editcardv}
\begin{editcardv}{Squid Red Idle}{squid-red-idle}{figures/tasks/squid-red-idle}
Animation, 3 frames \quad Input $58\times57$\par
\taskprompt{3 frames. Frame 1 keeps the original picture content (the canvas may be changed); frames 2 and 3 show the red squid monster idling, part of the sprite bobbing gently up and down for a breathing idle.}\par
\tasksource{Ninja Adventure Asset Pack by Pixel-boy and AAA (itch.io), Actor/Boss/SquidRed; frames 1--3 of \texttt{IdlePreview.gif}.}{CC0 1.0}
\end{editcardv}
\begin{editcardv}{Bride Turnaround}{char-turn-3view}{figures/tasks/char-turn-3view}
Multiview, 3 frames \quad Input $66\times119$\par
\taskprompt{The canvas already contains a pixel-art character. Keep the character's design, outfit and palette exactly the same and change only the viewing angle, producing three frames in this order: frame 1 a three-quarter view with the character turned 45 degrees to its left, frame 2 the opposite three-quarter view with the character turned 45 degrees to its right, frame 3 the rear view (seen from directly behind). Submit all 3 frames in order with finish([f1, f2, f3]).}\par
\tasksource{Hugging Face dataset \texttt{evilsocket/alucard-sprites}, images 0000000--0000003 (input 0000002).}{FAIR License v1.0.0}
\end{editcardv}
\begin{editcardv}{House Turnaround}{house-turn-2view}{figures/tasks/house-turn-2view}
Multiview, 2 frames \quad Input $112\times157$\par
\taskprompt{Rotate this house by 45 degrees and then by 90 degrees and draw both views, keeping the same 45-degree high-angle top-down view in both frames (a fixed elevated isometric camera: the camera height and pitch stay unchanged, so neither view becomes a flat eye-level elevation): the 45-degree frame shows the front door and front wall together with the window and wall on the right-hand side; the 90-degree frame shows only that right-hand wall and its window.}\par
\tasksource{Tiny Swords (Free Pack) by Pixel Frog (itch.io), Buildings/Yellow Buildings, \texttt{House1.png} (input), \texttt{House2.png} and \texttt{House3.png} (reference views).}{Tiny Swords free-pack license (no redistribution)}
\end{editcardv}
\end{tcbraster}

%% file: tables/table_slices_b2.tex
\begin{tabular}{@{}lcccccccc@{}}
\toprule
 & & & & & & \multicolumn{3}{c}{Leave-one-out} \\
\cmidrule(l){7-9}
Slice & Tasks & Score SD & Rater--rater & Rank SD & Judge--human & Judge & Raters & Below \\
\midrule
Characters             &  13 & 10.4 & 0.492 & 2.38 & 0.609 & 0.614 & 0.482--0.683 & 5 \\
Creatures              &  10 &  9.4 & 0.469 & 2.04 & 0.563 & 0.569 & 0.527--0.662 & 3 \\
Scenes, architecture   &  16 & 10.0 & 0.466 & 2.29 & 0.634 & 0.631 & 0.494--0.667 & 7 \\
Icons, interface       &   7 &  8.6 & 0.457 & 2.25 & 0.579 & 0.571 & 0.363--0.698 & 4 \\
Objects                &  21 &  7.1 & 0.448 & 2.06 & 0.518 & 0.508 & 0.381--0.662 & 2 \\
Effects                &   3 &  9.7 & 0.437 & 2.23 & 0.445 & 0.446 & 0.370--0.758 & 1 \\
Materials              &   5 &  8.0 & 0.263 & 1.61 & 0.460 & 0.453 & 0.174--0.549 & 7 \\
\midrule
Canvas $> 64$          &  14 & 11.5 & 0.504 & 2.28 & 0.697 & 0.696 & 0.559--0.694 & 10 \\
Canvas 33--64          &  33 &  8.8 & 0.478 & 2.29 & 0.573 & 0.569 & 0.480--0.683 & 3 \\
Canvas 17--32          &  18 &  7.7 & 0.417 & 1.97 & 0.494 & 0.494 & 0.372--0.662 & 2 \\
Canvas $\leq 16$       &  10 &  7.8 & 0.343 & 1.67 & 0.472 & 0.465 & 0.368--0.572 & 5 \\
\midrule
Canvas $\geq 64$       &  42 &  9.8 & 0.484 & 2.27 & 0.631 & 0.628 & 0.493--0.673 & 6 \\
Canvas 16--32          &  28 &  7.8 & 0.390 & 1.84 & 0.486 & 0.483 & 0.376--0.595 & 3 \\
\midrule
Instruction editing    &  13 & 14.1 & 0.474 & 2.03 & 0.529 & 0.527 & 0.404--0.643 & 4 \\
Animation              &  10 & 15.2 & 0.508 & 1.85 & 0.545 & 0.545 & 0.508--0.732 & 1 \\
Multiview              &   2 & 17.9 & 0.458 & 1.49 & 0.561 & 0.539 & 0.328--0.713 & 4 \\
\bottomrule
\end{tabular}

%% file: tables/table_cost_b2.tex
\begin{tabular}{@{}lccccccc@{}}
\toprule
Configuration & Calls & Writes & Input (k) & Cached & Output (k) & Minutes & Cost \\
\midrule
GPT-6 Astra (high)                       & 4.6 & 2.0 & 63 & 88\% & 2.3 & 1.9 & \$0.24 \\
Claude Opus 5 (low)                & 15.8 & 9.7 & 292 & 85\% & 7.5 & 7.3 & \$0.52 \\
Qwen3.8 Max                              & 15.9 & 11.1 & 310 & 91\% & 65.8 & 20.8 & \$0.52 \\
Gemini 3.8 Flash (high)                  & 27.0 & 13.2 & 1{,}143 & 79\% & 63.7 & 9.4 & \$0.49 \\
GPT-5.6 Sol (medium)                     & 5.7 & 2.2 & 82 & 89\% & 3.0 & 1.2 & \$0.13 \\
DeepSeek V4.1 Flash (high)               & 27.4 & 15.2 & 1{,}782 & 98\% & 76.9 & 8.4 & \$0.11 \\
GLM-5.3 Flash (max)                & 21.1 & 10.9 & 480 & 94\% & 197.8 & 31.1 & \$0.12 \\
Kimi K3 (max)                            & 9.9 & 5.8 & 126 & 93\% & 23.8 & 9.1 & \$0.42 \\
Gemini 3.1 Pro Preview (high)            & 8.2 & 4.6 & 159 & 44\% & 29.3 & 5.0 & \$0.55 \\
DeepSeek V4 Pro (high)                   & 14.3 & 7.6 & 584 & 97\% & 40.3 & 7.8 & \$0.21 \\
DeepSeek V4 Flash (high)           & 14.4 & 7.7 & 328 & 92\% & 41.9 & 7.8 & \$0.06 \\
Qwen3.5-397B-A17B                  & 10.2 & 4.7 & 183 & --- & 8.1 & 2.3 & \$0.14 \\
\bottomrule
\end{tabular}